\documentclass[journal,letterpaper,twoside,twocolumn]{IEEEtran}

\ifCLASSINFOpdf
  \usepackage[pdftex]{graphicx}
  \graphicspath{{./figures/}}
  \DeclareGraphicsExtensions{.eps,.pdf,.jpeg,.png,.tif}
\else
\fi

\usepackage{amsmath}
\usepackage{algorithm,algorithmic}
\usepackage{booktabs}
\usepackage{xcolor}
\usepackage{array}
\usepackage{multirow}
\usepackage{adjustbox}
\usepackage{amssymb}
\usepackage{float}
\usepackage{enumitem}
\usepackage{rotating}
\usepackage{physics}
\usepackage{mathtools}
\usepackage{pifont}
\usepackage{url}

\usepackage{hyperref}
\usepackage{cleveref}

\usepackage{xcolor}
\usepackage{changes}
\usepackage{bm}

\ifCLASSOPTIONcompsoc
  \usepackage[caption=false,font=normalsize,labelfont=sf,textfont=sf]{subfig}
\else
  \usepackage[caption=false,font=footnotesize]{subfig}
\fi

\newboolean{review}
\reviewfalse

\ifreview
\paperwidth=\dimexpr \paperwidth + 12cm\relax
\oddsidemargin=\dimexpr\oddsidemargin + 6cm\relax
\evensidemargin=\dimexpr\evensidemargin + 6cm\relax
\marginparwidth=\dimexpr \marginparwidth + 6cm\relax
\fi

\newcommand{\F}[1]{\textcolor{black}{#1}}

\begin{document}

\title{Multitask Learning for Earth Observation Data Classification with Hybrid Quantum Network}

\author{Fan Fan,
        Yilei Shi,~\IEEEmembership{Member,~IEEE},
        Tobias Guggemos,
        Xiao Xiang Zhu,~\IEEEmembership{Fellow,~IEEE}
\thanks{Fan Fan is with the Chair of Data Science in Earth Observation (SiPEO), Technical University of Munich (TUM), München, 80333, Germany, and also with the Remote Sensing Technology Institute (IMF), German Aerospace Center (DLR), Weßling, 82234, Germany}
\thanks{Yilei Shi is with the School of Engineering and Design, TUM, München, 80333, Germany}
\thanks{Tobias Guggemos is with IMF, DLR, 82234 Weßling, Germany, and also with the  University of Vienna, Faculty of Physics, Vienna Center for Quantum Science and Technology (VCQ), 1090 Vienna, Austria}
\thanks{Xiao Xiang Zhu is with SiPEO, TUM, 80333 Munich, Germany, and also with Munich Center for Machine Learning, 80333 Munich, Germany}
\thanks{\F{Our trained models and code are available at: https://github.com/zhu-xlab/MLTQNN.}}
}

% \markboth{submit to IEEE Transactions on Geoscience and Remote Sensing, 2025}%
% {}
\maketitle

\begin{abstract}
Quantum machine learning (QML) has gained increasing attention as a potential solution to address the challenges of computation requirements in the future. Earth observation (EO) has entered the era of Big Data, and the computational demands for effectively analyzing large EO data with complex deep learning models have become a bottleneck. Motivated by this, we aim to leverage quantum computing for EO data classification and explore its advantages despite the current limitations of quantum devices. This paper presents a hybrid model that incorporates multitask learning to assist efficient data encoding and employs a location weight module with quantum convolution operations to extract valid features for classification. The validity of our proposed model was evaluated using multiple EO benchmarks. Additionally, we experimentally explored the generalizability of our model and investigated the factors contributing to its advantage, highlighting the potential of QML in EO data analysis.
\end{abstract}

\begin{IEEEkeywords}
Quantum Machine Learning, Quantum Circuit, Image Classification, Multitask Learning, Generalizability, Earth Observation, Remote Sensing
\end{IEEEkeywords}

\section{Introduction}

Earth observation (EO) data classification aims to extract critical information from remote sensing imagery,  offering valuable insights for a wide range of applications, such as urban planning, resource management, environmental monitoring, and more. To analyze EO data efficiently and accurately, the exploration of machine learning approaches has gained significant attention, as these methods facilitate the automatic interpretation of large volumes of EO data. To date, many contributions have been made in this regard, as demonstrated in studies such as \cite{zhu2017deep, zhu2021deep}. 

However, to address the complexities of EO data analysis with machine learning techniques, researchers have developed advanced model architectures and training strategies. For instance, Wang et al.~\cite{wang2024decoupling} proposed a method to decouple common and unique representations across different modalities, enhancing model performance. However, such complex techniques come at the cost of significant computational power requirements, which have become a bottleneck. To tackle this challenge, quantum computing has gained significant attention for its potential to accelerate computational tasks despite the current limitations of quantum devices, including their lack of full fault tolerance and limited qubit support.

Quantum machine learning (QML) integrates quantum computing and machine learning, and various contributions have been made to explore its potential across different tasks. Wei et al. \cite{wei2022quantum} proposed a quantum convolutional neural network for image recognition tasks, which shows exponential speed-ups over their classical counterparts in gate complexity. De et al. \cite{de2024quantum} proposed a quantum latent diffusion model for generation tasks, highlighting several advantages of quantum computing, such as faster convergence and fewer trainable parameters.  

Despite such existing contributions, there are still many challenges in this field. For instance, data encoding is a critical and challenging step in QML, as it directly impacts both the validity of QML for data analysis and the overall efficiency of the model. This challenge is particularly significant in EO data analysis due to the rich spectral and spatial information contained in EO imagery. To address this, many proposed methods adopt a hybrid framework, leveraging quantum algorithms explicitly for latent feature transformation \cite{zollner2024satellite} or local low-level feature transformation \cite{sebastianelli2024quanv4eo}. 

Furthermore, efficiency might not be the only benefit of using QML for EO data analysis. Transferability and generalizability are also important factors. Data scarcity and variability are common issues when addressing global challenges in the EO domain. High transferability and generalizability ensure that models trained with limited labeled data or data from different domains remain adaptable and robust across diverse applications. In the study \cite{wang2023hybrid}, Wang et al. introduced a quantum framework for learning transferable visual representation, where the quantum component enhances both transferability and generalization. However, whether quantum computing can improve the transferability and generalizability of EO data analysis is still uncertain.

\F{To address these challenges and fill the research gaps, our prior work \cite{fan2025hybrid} introduced the SEQNN model that exploits a classical multilayer perceptron to assist efficient image data encoding. However, achieving effective feature reduction requires a large number of training samples to ensure robust performance, which could restrict its practicality and potential, especially when handling large EO images with limited training samples. This key limitation motivates the present study, in which we propose a multitask-based hybrid quantum neural network (MLTQNN). Specifically, this model leverages a reconstruction task learning to reduce features for efficient and valid quantum data encoding. For classification, we introduce a location weight module alongside quantum convolution operators to extract critical features. In addition, this study also further explores the generalizability of QML models in EO data classification and investigates the underlying factors that contribute to it. The performance and validity of our model for EO data classification have been thoroughly evaluated using multiple EO benchmarks under different settings, providing meaningful insights into the behavior and generalization capability of QML models in EO data analysis.}

In summary, the main contributions of this work include:

\begin{enumerate}
  \item \F{We proposed a hybrid neural network~(MLTQNN) that incorporates a multitask learning approach to assist efficient feature encoding by significantly reducing the number of encoded features through an auxiliary image reconstruction task for effective and efficient feature extraction and achieve accurate EO imagery classification;}  
  \item We introduced a location weight module to enhance feature extraction along with quantum convolution operations;
  \item \F{We studied the generalizability advantages of our model and its underlying reasons from different perspectives, highlighting the potential of QML in EO data analysis in this regard;}
\end{enumerate}

The paper is organized as follows: Section~\ref{relatedwork} overviews related work. Section~\ref{methodology} introduces our proposed hybrid model. Section~\ref{experiment} presents the experiments and their results to evaluate our model. Section~\ref{discussion} discusses the generalizability advantages of our hybrid model and analyzes the underlying reasons. Finally, Section~\ref{conclusion} concludes the paper and outlines future research directions.   

\section{Related Work}\label{relatedwork}

This section briefly overviews relevant studies, highlighting multitask learning strategies in the EO domain, contributions to the generalizability of quantum machine learning models, and recent explorations of QML for EO data classification.  

\subsection{Multitask Learning for EO Data Analysis}

Multitask learning has been shown to leverage information from multiple related tasks to enhance the generalization performance across all the tasks\cite{zhang2021survey}. Numerical studies have been conducted to apply this approach in EO data analysis. Zheng et al.~\cite{zheng2021generalized} proposed a framework for jointly learning shared features from multiple small-scale EO data alongside task-specific features to achieve optimal results. Carvalho et al.~\cite{carvalho2019multitask} introduced a network learning semantics and local heights together, enabling each task to benefit from it. Liu et al.~\cite{liu2024mlcnet} proposed a model that can enhance targeted emphasis on features at different levels through multitask learning constraints. Leiva-Murillo et al.~\cite{leiva2012multitask} aimed to alleviate the dataset shift for EO data classification by imposing cross-information in the classifiers through matrix regularization. Li et al.~\cite{li2024mt} integrated a hierarchical classification network and a classification-assisted attention-based regression network to retrieve multiple cloud properties. Qiu et al.~\cite{qiu2020multitask} introduced multitask learning to human settlement extent regression and local climate zone classification and exploited the former output as a prior to improve classification accuracy. Liu et al.\cite{liu2020few} came up with a multitask learning method to simultaneously conduct classification and reconstruction tasks using hyperspectral images.

\subsection{Quantum Machine Learning for EO Data Classification}

To date, an increasing amount of contributions has been made regarding applying QML for EO data classification despite the current constraints of quantum devices. Some studies have focused on using quantum annealers for classification tasks, such as \cite{delilbasic2023single, zardini2024local}. In addition, quantum circuit-based models have been explored to classify EO data. However, due to the complexity and richness of EO data, hybrid quantum deep learning models have attracted significant attention. For instance, Rodriguez-Grasa et al.~\cite{rodriguez2024satellite} utilized the U-Net Encoder to reduce the dimensionality of EO data and neural quantum kernels for final prediction. Miroszewski et al.~\cite{miroszewski2023detecting} applied simple linear iterative clustering for data reduction, followed by their proposed quantum-kernel support vector machines for multispectral image analysis. Zollner et al.~\cite{zollner2024satellite} compared different dimensionality reduction methods for EO data as part of QML models for classification purposes. 
Miller et al.~\cite{miller2023quantum} presented a hybrid quantum-classical model that combines VGG or RNN architectures with a quantum circuit for optical and Synthetic Aperture Radar (SAR) classification. 

Besides using classical algorithms for preprocessing EO images followed by quantum components, some studies rely on quantum circuits to directly extract features from EO images, with classical algorithms employed for final classification. Examples of such models can be found in ~\cite{sebastianelli2024quanv4eo, fan2023hybrid}.

Furthermore, Papa et al.\cite{papa2024impact} explored the benefits of hybrid quantum attention-based models for enhancing classification performance and stability in EO data analysis. Miroszewski et al.~\cite{miroszewski2023quantum} outlined the potentials and challenges of applying QML in the EO domain. 

\subsection{Generalizability in Quantum Machine Learning}

Generalizability is a critical aspect of machine learning models, reflecting their ability to perform on unseen data. Numerous studies have explored the generalization bounds of QML models from various perspectives. For instance, studies such as \cite{caro2022generalization, du2023problem, gibbs2024dynamical} suggest that a QML model’s generalizability is influenced by the number of training samples, a finding that is consistent with classical machine learning models. 

\begin{figure*}[ht]
    \centering
    \includegraphics[width=.85\linewidth]{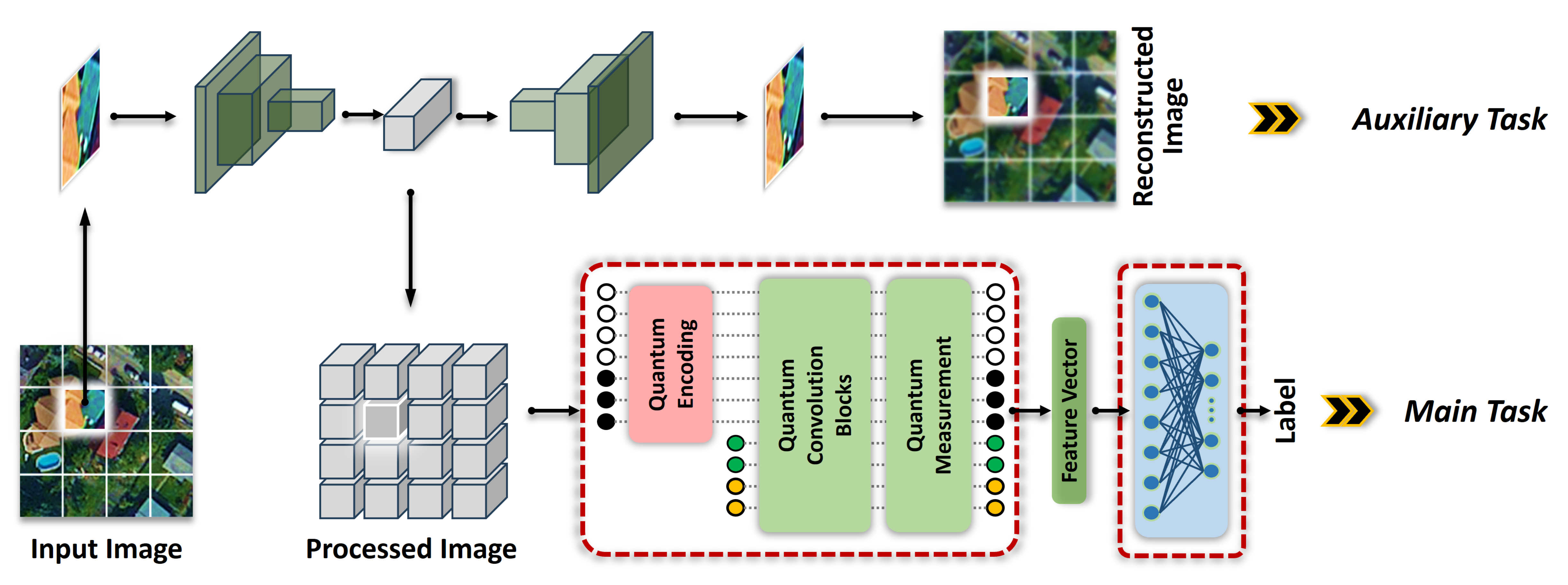}
    \caption{Overall architecture of the proposed model: image classification as the main task and image reconstruction as the auxiliary task}
    \label{fig: overall}
\end{figure*}

In addition, several other parameters of QML models have been linked to generalization bounds. Specifically, besides the size of the training data, Caro et al.~\cite{caro2023out} showed that generalization depends on the number of trainable quantum gates. Qi et al.~\cite{qi2023theoretical} demonstrated a connection between generalization and the Hilbert space dimension. Banchi et al.~\cite{banchi2021generalization} explored generalization in QML through quantum information theory, suggesting that the generalization error is connected to the Rényi mutual information between the quantum state space and the classical parameter space. Caro et al.~\cite{caro2021encoding} emphasized the critical role of data-encoding strategies in the generalizability of QML by obtaining the generalization bounds based on the Rademacher complexity and the metric entropy from statistical learning theory. Gyurik et al.~\cite{gyurik2023structural} focused on the relationship between the applied observable in QML and its generalization performance based on the Vapnik–Chervonenkis dimension. 

Furthermore, Khanal et al.~\cite{khanal2024generalization} presented a comprehensive summary of existing studies on generalization error bounds in the NISQ era. Interested readers may refer to this work for a detailed overview.

\section{Methodology}\label{methodology}
% To address these challenges and fill the research gaps, our prior work \cite{fan2025hybrid} introduced the SEQNN model that exploits a classical multilayer perceptron to assist efficient EO data encoding. 

\F{In this paper, we introduce a multitask-based hybrid quantum neural network (MLTQNN) that integrates image classification as the primary task and image reconstruction as a supporting auxiliary task. As depicted in \Cref{fig: overall}, the image reconstruction task is realized through a convolutional autoencoder, while the image classification task is performed by the hybrid quantum neural network. The motivation is to leverage the reconstruction task for efficient and valid quantum data encoding, thereby improving EO data classification, particularly when handling large EO images with limited training samples. A detailed description of each task is provided below.}

\subsection{Main Task Learning}\label{classificationtask}
For the main task, our proposed model employs a hybrid quantum algorithm to classify EO images, leveraging quantum properties to enhance feature extraction. As shown in \Cref{fig: overall}, this task consists of two components: a quantum algorithm for feature extraction and a classical algorithm for final prediction using the feature vector obtained from quantum computing. Specifically, the process comprises four steps, which we detail in the following sections.

\begin{figure*}[ht]
    \centering
    \includegraphics[width=.8\linewidth]{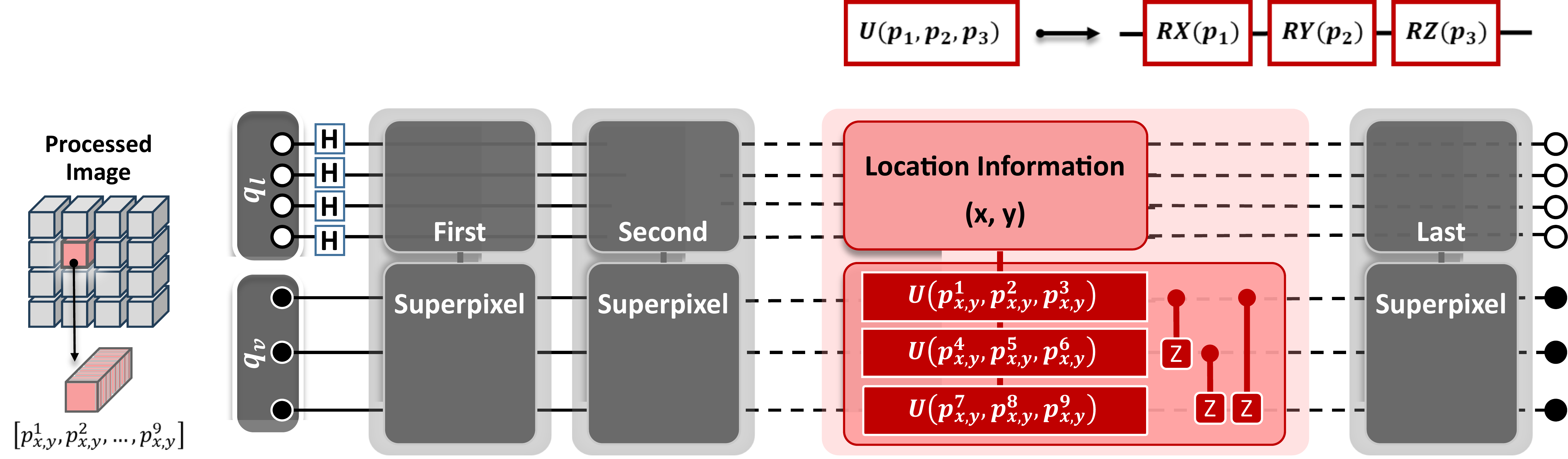}
    \caption{\F{The Quantum Circuit Example for Encoding: 1) $p_{x,y}^{i}$ indicates $i^{th}$ feature value of the superpixel $p_{x,y}$ 2) color indicates the quantum register: white qubits for $q_{l}$ and black qubits for $q_{v}$ 3) H denotes the Hadamard gate, while RX, RY, and RZ indicate rotation gates.}}
    \label{fig: encoding_circuit}
\end{figure*}

\subsubsection{Data Encoding} Representing classical data as quantum states is a critical step in quantum machine learning. For EO data, this process is particularly challenging due to the richness of its spectral and spatial information, requiring efficient methods to minimize quantum resource usage. In our proposed model, the processed image derived from the auxiliary task is encoded into the quantum state. This approach reduces the quantum resources needed for the encoding process, effectively addressing the challenges associated with encoding complex EO data.

\F{Following the encoding strategy in \cite{fan2025hybrid}, the processed image described in \Cref{reconstructiontask} is transformed into a quantum state via a quantum circuit, and \Cref{fig: encoding_circuit} illustrates an example for encoding the processed image with $4\times4$ superpixels, each containing $9$ feature values.} As indicated in the figure, there are two quantum registers, distinguished by white and black qubits, respectively. The first register, $q_{l}$, is responsible for the locations of the superpixels, while the second register, $q_{v}$, encodes their feature values. \F{To be more specific, we first encode all the coordinate pairs of the superpixels in the processed image into the quantum domain by applying Hadamard gates to $q_{l}$, thereby representing all the location information using a superposition with equal amplitude. Then, to encode the features of a superpixel, represented by $[p_{x,y}^{1}, p_{x,y}^{2}, \dots, p_{x,y}^{9}]$, where $(x,y)$ denotes its coordinate, we use three rotation gates—RX, RY, and RZ—to each qubit in $q_{v}$, with each gate encoding one feature value.} \Cref{encoding_featurevalue} indicates the state of qubit $q^{n}_{v}$ in $\ket{q_v}$ after the rotation gates. 

\begin{equation} \label{encoding_featurevalue}
\begin{aligned}
    \ket{q^{n}_{v}} &= e^{-i\frac{p^{3n}_{x,y}}{2} Z} e^{-i\frac{p^{3n-1}_{x,y}}{2} Y} e^{-i\frac{p^{3n-2}_{x,y}}{2} X}\ket{0}, n \in [1,2,3].
\end{aligned}
\end{equation}

\F{To associate the location of each superpixel with its feature values, we entangle these two registers using controlled rotation gates on $q_{v}$, where the control is determined by the state of $q_{l}$.} Each superpixel is encoded sequentially, resulting in the following quantum state for a processed image with $2^n\times2^n$ superpixels:

\begin{equation}\label{encoding_state1}
    \ket{\Psi_{image}}= \frac{1}{2^{n}}\sum_{x=1}^{2^n}\sum_{y=1}^{2^n}\ket{l_{x,y}}\ket{p_{x,y}}
\end{equation} in which $\ket{l_{x,y}}$ denotes the $(x,y)$-coordinate of a superpixel in binary string form, represented as $\ket{x_{n} \dots x_{1} y_{n} \dots y_{1}}$, while $\ket{p_{x,y}}$ indicates its feature values. \F{Meanwhile, a CZ gate is also applied between each pair of qubits in $q_{v}$ for each superpixel, and the resulting quantum state can be expressed as:}

\begin{equation}\label{encoding_state2}
\begin{aligned}
    \ket{\Psi_{image}^{\ast}}= \frac{1}{2^{n}}\sum_{x=1}^{2^n}\sum_{y=1}^{2^n}\ket{l_{x,y}}\ket{p_{x,y}^{\ast}}
\end{aligned}    
\end{equation} in which $\ket{l_{x,y}}$ denotes the location of a superpixel and $\ket{p_{x,y}^{\ast}}$ indicates the resulting state of the feature values after applying the CZ gates. \Cref{encoding_state3} represents the transformation from $\ket{p_{x,y}}$ to $\ket{p_{x,y}^{\ast}}$.

\begin{equation}\label{encoding_state3}
\begin{aligned}
   \ket{p_{x,y}} =  & a\ket{000} + b\ket{001} + c\ket{010} + d\ket{011}  + \\ & e\ket{100} + f\ket{101} + g\ket{110} + h\ket{111} \\
   \implies \ket{p_{x,y}^{\ast}} = & a\ket{000} + b\ket{001} + c\ket{010} - d\ket{011}  + \\ & e\ket{100} - f\ket{101} - g\ket{110} - h\ket{111} \\
\end{aligned}    
\end{equation} 

\begin{figure*}[ht]
    \centering
    \includegraphics[width=.75\linewidth]{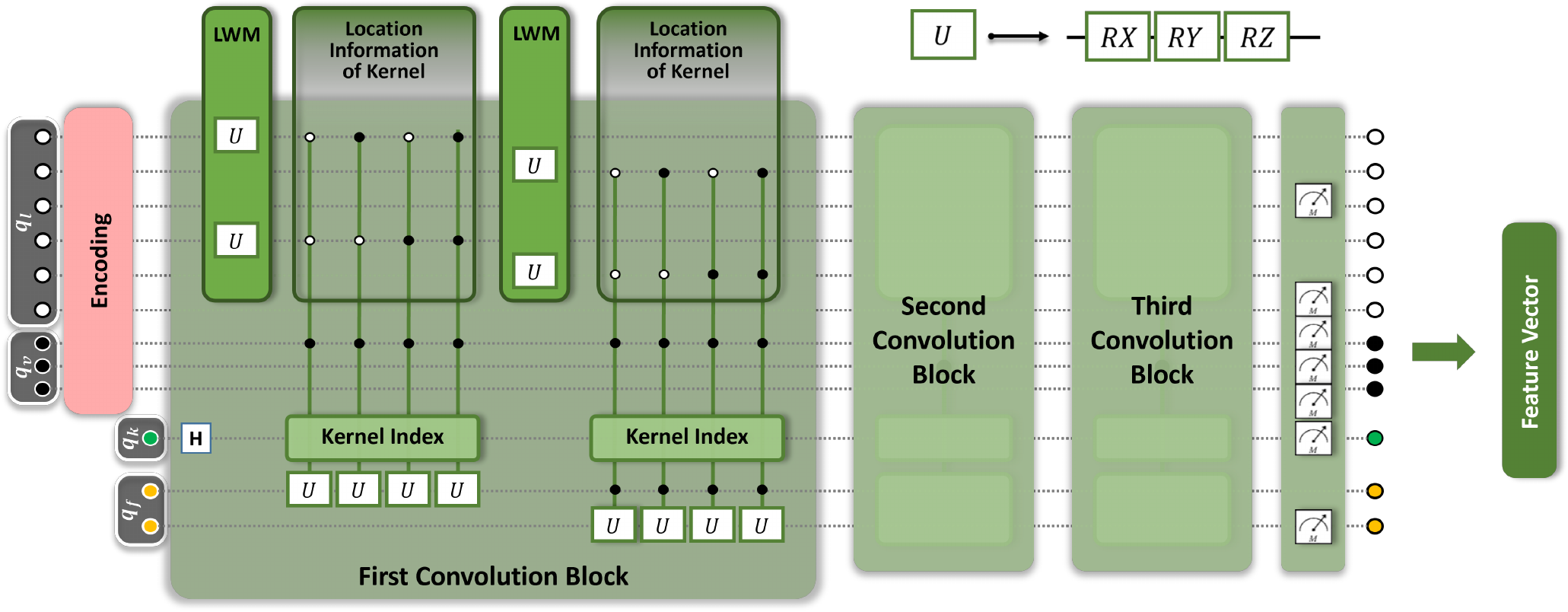}
    \caption{\F{The Quantum Circuit Example for Feature Extraction: 1) color indicates the quantum register: white qubits for $q_{l}$, black qubits for $q_{v}$, green qubits for $q_{k}$ and yellow qubits for $q_{f}$ 2) U denotes the gate unit consisting of RX,RY and RZ gates 3) dot markers indicate the controlled state: white dots for $\ket{0}$ and black dots for $\ket{1}$ 4) LWM stands for Location Weight Module, newly introduced to enhance feature extraction}}
    \label{fig: pqc_conv_circuit}
\end{figure*}

After encoding all the superpixels in this manner, the input image is represented as a quantum state, enabling further feature extraction in the quantum domain.

\subsubsection{Feature Extraction} To extract features from EO data, the convolution operation plays an important role due to its effectiveness and adaptability in various deep learning models. \F{Our proposed model also relies on quantum convolution operations to extract meaningful features from the encoded quantum states, and its architecture is rooted in our previously proposed SEQNN model \cite{fan2025hybrid}, with the quantum convolution blocks preserved. In this work, we additionally introduce a Location Weight Module (LWM) within the convolution blocks to further enhance the extracted features. For completeness, we describe the implementation of both the quantum convolution operations and the newly introduced LWM module below. }

\F{The implementation is illustrated in \Cref{fig: pqc_conv_circuit}. Besides the quantum registers $q_{l}$ and $q_{v}$, two additional quantum registers are involved: $q_{k}$ represents the indices of the kernels and the resulting feature maps, and $q_{f}$ stores the values of the corresponding generated feature maps. As shown in the figure, the circuit consists of several convolution blocks, each corresponding to a qubit in $q_{v}$. Within each block, multiple kernels can be applied for convolution operations, generating feature maps for further analysis. Specifically, to exploit $2^k$ kernels, we need $k$ qubits in $q_{k}$, and its state indicates the index of the applied kernel as well as the corresponding generated feature map.} The number of qubits in $q_{f}$ is determined by the size of the feature maps, with each qubit encoding the values of the feature maps produced by the applied convolution layers. As indicated in the figure, the controlled gate units, consisting of RX, RY and RZ gates, are used as kernels for convolution operations, with their parameters being optimized during the training process. 

\begin{figure}[t]
    \centering
    \includegraphics[width=.75\linewidth]{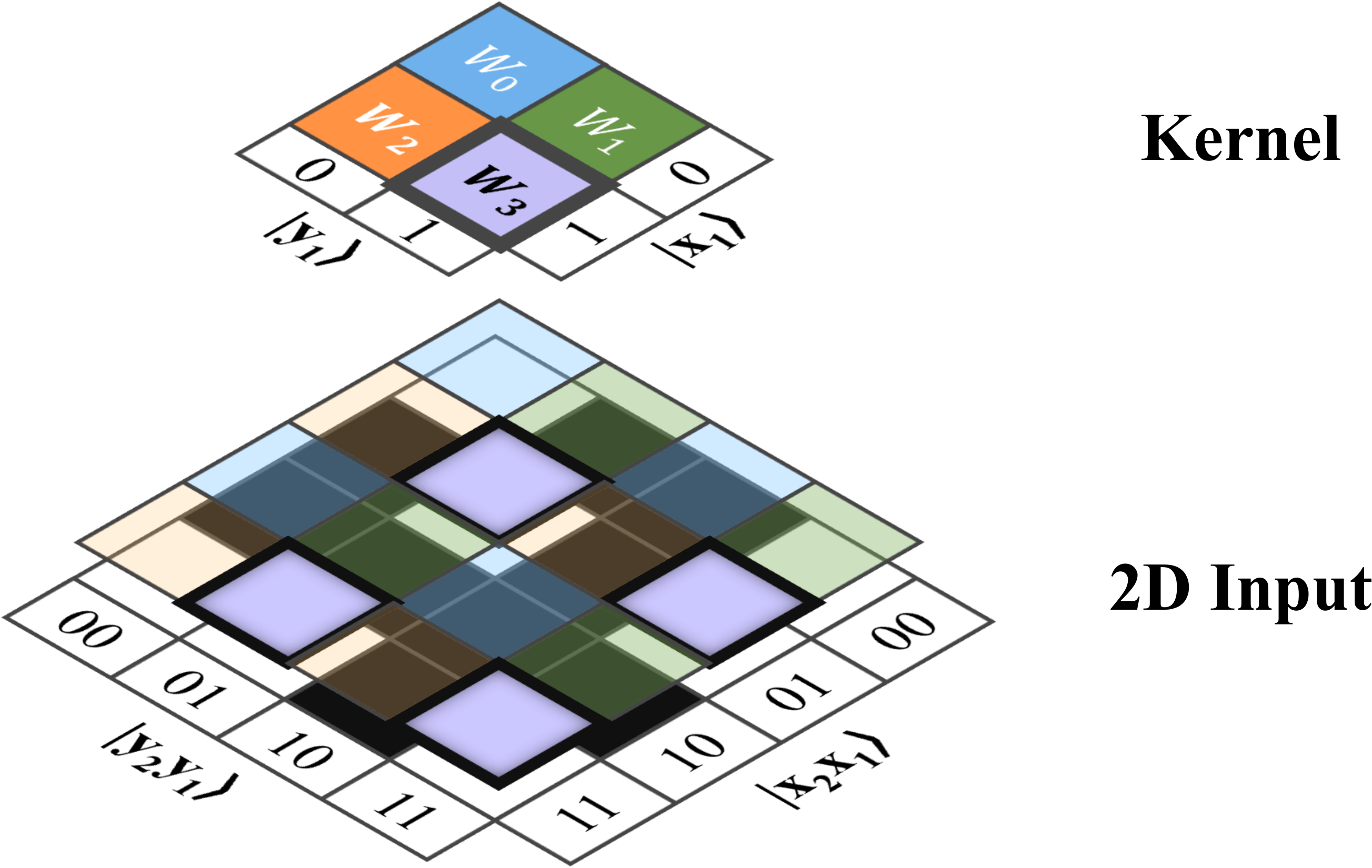}
    \caption{\F{Illustration of the quantum convolutional operation using a $2\times2$-sized kernel with weights from $W_{0}$ to $W_{3}$: the quantum states $\ket{x_{2}x_{1}y_{2}y_{1}}$ in $q_{l}$ represent the locations of a $4\times4$-sized image, and the state $\ket{x_{1}y_{1}}$ can be used to identify the pixels transformed with the same weights. The visualization highlights $W_3$ as an example, where the state $\ket{11}$ selects its corresponding pixels for transformation}}
    \label{fig: convolutionOp}
\end{figure}

To provide more details on the quantum convolution operation, \Cref{fig: convolutionOp} illustrates its principle. For this operation, the convolution stride is set to match the kernel size. As shown in the figure, the pixels from a $4\times4$-sized image with the same color are transformed with the same weight. By identifying the state $\ket{x_{1}y_{1}}$ in $q_l$, we can determine all the pixels that should be transformed with the same weight. Together with the register $q_v$, which encodes the pixel values, we can perform the element-wise operation in the quantum domain using controlled gates. In our proposed model, we use a gate unit, consisting of RX, RY and RZ gates, controlled by the state of two qubits in register $q_l$ and the register $q_v$. The obtained quantum state after the element-wise operation can be indicated as:

\begin{equation}\label{kernel_computation1}
    \ket{\Psi_{f_{1}}}= \frac{1}{2^{n}}\sum_{x=1}^{2^n}\sum_{y=1}^{2^n}\ket{l_{x,y}}\ket{p_{x,y}^{\ast}}\ket{w_{x,y}}
\end{equation} where $\ket{l_{x,y}}$ denotes the location of a superpixel, $\ket{p_{x,y}^{\ast}}$ represents its encoded feature values, and $\ket{w_{x,y}}$ denotes the weights of the kernels. 

\F{To further enhance feature extraction in this step, we additionally introduce and integrate a Location Weight Module (LMW) in the convolution blocks. Specifically, this module applies a trainable gate unit, including RX, RY, and RZ gates, to the corresponding controlled qubits from $q_l$ for the quantum convolution operations. As a result, the amplitude for $\ket{l_{x,y}}$ can be optimized based on the task, rather than remaining constant. The quantum state after applying the Location Weight Module is represented as $\ket{l_{x,y}^{\ast}}$, and the corresponding state after the convolution operations can be indicated as: }

\begin{equation}\label{kernel_computation2}
    \ket{\Psi_{f_{1}}^{\ast}}= \frac{1}{2^{n}}\sum_{x=1}^{2^n}\sum_{y=1}^{2^n}\ket{l_{x,y}^{\ast}}\ket{p_{x,y}^{\ast}}\ket{w_{x,y}}
\end{equation} 

To obtain the feature map, we integrate the outputs of the element-wise product from each region. The generated feature map $\ket{\Psi_{f_{2}}^{\ast}}$ can be indicated as:

\begin{equation} \label{kernel_computation3}
    \ket{\Psi_{f_{2}}^{\ast}}= \frac{2}{2^{n}}\sum_{x'=1}^{2^{n-1}}\sum_{y'=1}^{2^{n-1}}\ket{l_{x',y'}}\ket{f_{x',y'}}
\end{equation} in which $\ket{l_{x',y'}}$ indicates the location information of the feature map, and $\ket{f_{x',y'}}$ encodes its value which can be represented as:

\begin{equation} \label{kernel_computation4}
    \ket{f_{x',y'}} = \frac{1}{2}\sum_{x_{1}=0}^{1}\sum_{y_{1}=0}^{1}\ket{l_{x'x_{1},y'y_{1}}^{\ast}}\ket{p_{x'x_{1},y'y_{1}}^{\ast}}\ket{w_{x'x_{1},y'y_{1}}}
\end{equation}

When multiple kernels are applied, we first apply Hadamard gates to the qubits in $q_k$. Then, additional controllers from $q_k$ are introduced to the controlled gates, generating the corresponding feature maps. These feature maps can then be used as input for the next quantum convolution layer. For feature extraction, multiple convolution layers can be applied, forming a convolution block associated with each qubit in $q_{v}$. By utilizing several such blocks, the target feature maps $\ket{\Psi_{target}}$ are obtained for further classification

\subsubsection{Measurement}

To extract a feature vector from the generated feature maps $\ket{\Psi_{target}}$, we define a set of operators $M$. The corresponding expectation values with respect to $\ket{\Psi_{target}}$ are then used to construct the feature vector for final classification. 

To clarify, as shown in \Cref{fig: pqc_conv_circuit}, only the qubits that encode the target feature maps are measured. In the example provided, these include two qubits $q_l^{t_x}$ and $q_l^{t_y}$ from $q_l$, all qubits from $q_v$ and $q_k$, as well as the last qubit $q_f^{t}$ from $q_f$. The operators $M$ are constructed using identity and Pauli-X operators. The following equation indicates the generation of the operator set $M$. 

\begin{equation}\label{measure_operator}
\begin{aligned}
  M = & (I \pm X_{q_l^{t_x}}) \otimes (I \pm X_{q_l^{t_y}}) \otimes (I - X_{q_f^{t}}) \\ & \otimes \prod_{q_k^{i} \in q_k} (I \pm X_{q_k^{i}})  \otimes \prod_{q_v^{i} \in q_v} (I \pm X_{q_v^{i}}) 
\end{aligned} 
\end{equation}

For each operator $M_i$, an expectation value is computed regarding the feature maps $\ket{\Psi_{target}}$ following \Cref{measure_expectation}, and the result is treated as one feature value in the generated feature vector.

\begin{equation}\label{measure_expectation}
  E(M_i) = \bra{\Psi_{target}} M_i \ket{\Psi_{target}}
\end{equation}

\subsubsection{Prediction} With the obtained feature vector, we apply a classical dense layer to perform the final classification, where each neuron encodes one value of the feature vector, and all neurons are fully connected. For the final output, a softmax activation function is used to produce a probability distribution indicating the label of the input image.

\begin{figure*}[ht]
    \centering
    \includegraphics[width=.8\linewidth]{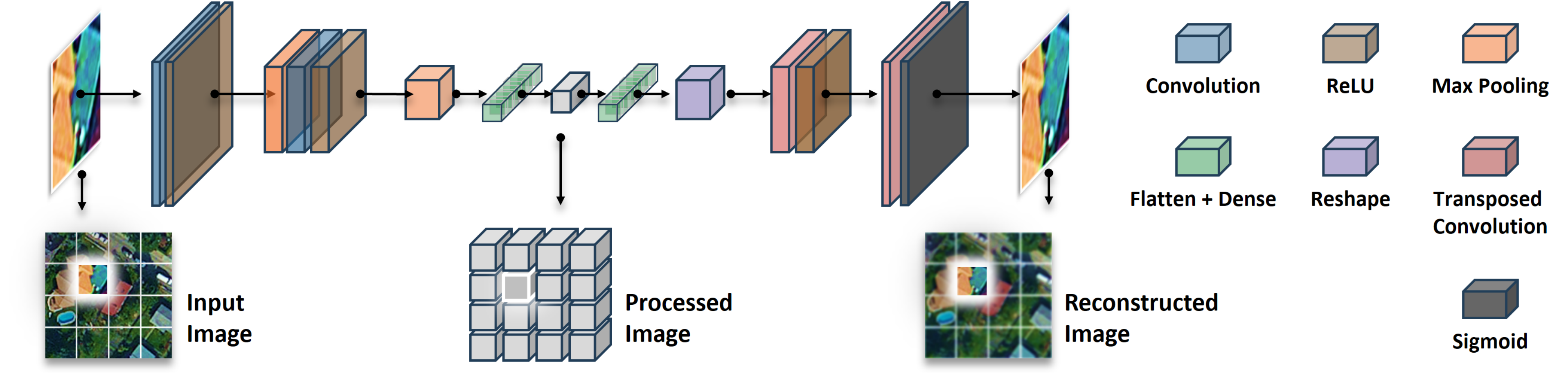}
    \caption{model structure for the reconstruction task}
    \label{fig: reconstruction_model}
\end{figure*}

\begin{table*}[th]
\centering
\caption{\F{Required Quantum Resources: \\ N: input image's spatial dimension; P: patch size; E: feature number \\ M: quantum convolution layer numbers; K: kernel numbers}}
\begin{tabular}{ccccc}
\toprule
\multirow{2}{*}{} & \multirow{2}{*}{Qubits} & \multicolumn{3}{c}{Quantum Gates} \\ \cmidrule{3-5} 
 &  & Gate Unit & Hadamard gates & CZ gates \\ \midrule
Data Encoding & $2\log_2(N/P) + E/3$ & ${EN^2}/{3P^2}$ & $2\log_2(N/P)$ & ${N^2E(E-3)}/{18P^2}$ \\ \midrule
Feature Extraction & $M+\log_2(K)$ & $(4MKE+2ME+2E)/3$ & $\log_2(K)$ & \textbackslash{} \\ \bottomrule
\end{tabular}
\label{table: quantum_resources}
\end{table*}

\subsection{Auxiliary Task Learning}\label{reconstructiontask}

The auxiliary task aims to generate valid representations of the superpixels through image reconstruction ensuring that the important information of the input image is efficiently encoded into the quantum state while minimizing the use of quantum resources. \Cref{fig: reconstruction_model} illustrates the model structure for this task, which follows an encoder-decoder architecture. 

As shown in the figure, the input image is first split into small patches, with each patch corresponding to a superpixel. For each patch, an encoder composed of convolutional layers, pooling layers with the ReLu activation function, followed by a dense layer, is used to extract a small set of features. These features are then treated as the feature values for the corresponding superpixel. Then, a decoder consisting of a dense layer with transposed convolution layers is employed to reconstruct the provided patch using the extracted feature values. Once all the patches are reconstructed, the full image is obtained.

During this task, the obtained superpixels form a processed image that retains the original location information. This processed image serves as the input for the main task, as previously introduced.

\subsection{Multi-objective Loss}\label{loss}

To train our model, we optimize two tasks, image classification and image reconstruction, in a supervised manner. As illustrated in \Cref{fig: overall}, for an arbitrary normalized EO image $X_i$ along with its label $Y_i$, our model yields two outputs: the reconstructed image $\hat{X}_i$ from the image reconstruction task and the predicted label $\hat{Y}_i$ from the image classification task. Accordingly, two loss functions are utilized to optimize our model. 

Specifically, for image classification, we adopt the cross-entropy loss function. When there are $c$ categories, we transform $Y_i$ and $\hat{Y}_i$ into $[Y_{i}^{1},\dots, Y_{i}^{c}]$, $[\hat{Y}_{i}^{1},\dots, \hat{Y}_{i}^{c}]$, respectively, using one-hot encoding technique. The loss value $L_{ce}$ for this task with $N$ pairs of samples can be calculated as:

\begin{equation}\label{cross_entropy}
    L_{ce} = - \frac{1}{N} \sum_{i=1}^N \sum_{c=1}^C Y_{i}^{c} \log(\hat{Y}_{i}^{c})
\end{equation} 

Regarding image reconstruction, we use the mean squared error (MSE) loss function, which calculates the difference between the input image $X_i$ and the reconstructed image $\hat{X_i}$ as follows:  

\begin{equation}\label{MSE}
    L_{mse} = \frac{1}{N} \sum_{i=1}^N | X_i - \hat{X}_i |^2 
\end{equation}

However, $L_{ce}$ and $L_{mse}$ operate on different scales during the training process. To balance the training between the main task (image classification) and the auxiliary task (image reconstruction), we introduce a hyperparameter $\alpha$, which is discussed in detail in \Cref{Hyperparameter}. The overall loss is then computed as:

\begin{equation}\label{overallLoss}
    Loss = L_{ce} + \alpha L_{mse}
\end{equation}

All trainable parameters in our model are optimized and updated during backpropagation using gradients with respect to $Loss$, applying the Adam algorithm~\cite{kingma2014adam}. For gradient calculations of trainable parameters within the quantum circuit, we employ the adjoint method~\cite{luo2020yao}.

\subsection{Quantum Resource Requirement}

\begin{table*}[t]
\centering
\caption{Experimental Data for Performance Evaluation}
\setlength{\tabcolsep}{10pt}
\begin{tabular}{ccccccc}
\toprule
Data & Classes & Input Size & Training & Validation & Test & Patch Size \\ \midrule
SAT-6 & 6  & $32\times32\times4$ & $4200$ & $1200$ & $1200$ & $4\times4\times4$ \\ 
LCZ42 & 5  & $32\times32\times4$ & $6000$ & $2000$ & $2000$ & $4\times4\times4$ \\ 
EuroSAT & 5  & $64\times64\times3$ & $3000$ & $1000$ & $1000$ & $8\times8\times3$ \\ 
PatternNet &  4 & $256\times256\times3$ & $1920$ & $640$ & $640$ & $32\times32\times3$ \\
\bottomrule
\end{tabular}
\label{table: experimental_data}
\end{table*}

% \begin{table*}[ht]
% \centering
% \caption{\F{Experimental Model Structure Comparison}}
% \setlength{\tabcolsep}{10pt}
% \begin{tabular}{cccccc}
% \toprule
% \multirow{3.5}{*}{Model} & \multicolumn{2}{c}{Encoding} & \multicolumn{2}{c}{Feature Extraction} & \multirow{3.5}{*}{Classifier} \\ \cmidrule{2-5}
%  & \begin{tabular}[c]{@{}c@{}}Superpixel \\ Preprocessing\end{tabular} & \begin{tabular}[c]{@{}c@{}}Quantum \\ Encoding\end{tabular} & \begin{tabular}[c]{@{}c@{}}Convolution \\ Operation\end{tabular} & \begin{tabular}[c]{@{}c@{}}Location Weight \\ Module\end{tabular} &  \\ \midrule
% ours & Auxiliary Learning Task & Sequential Encoding & Quantum &  \checkmark & Dense \& Softmax \\ 
% MLTCNN & Auxiliary Learning Task & \xmark & Classical & \xmark & Dense \& Softmax \\
% SEQNN\cite{fan2025hybrid} & Non-linear Projection & Sequential Encoding & Quantum & \xmark & Dense \& Softmax \\
% CNN & \xmark & \xmark & Classical & \xmark & Dense \& Softmax \\ \bottomrule
% \end{tabular}
% \label{table: experimental_model}
% \end{table*}

\F{The quantum resources required to construct a QML model are critical in the NISQ era. Thus, we also discuss and analyze the number of qubits and quantum gates needed in our model for classification tasks, summarized in \Cref{table: quantum_resources}. As shown in the table, the required quantum resources depend on four key parameters: for the auxiliary task, the patch size $P$ and the feature number $E$; and for the main task, the number of quantum convolution layers $M$ and the number of kernels $K$ in each layer. These parameters can be adjusted according to the task requirements and the available computational resources.}

\section{Experiments}\label{experiment}

\subsection{Experimental Datasets}

To validate our model, we utilized four EO benchmarks: SAT-6\cite{basu2015deepsat}, LCZ42\cite{zhu2019so2sat}, EuroSAT\cite{helber2019eurosat}, and PatternNet\cite{zhou2018patternnet}. To efficiently conduct experiments for evaluation, we reduced the size of these datasets by focusing on specific categories or limiting the number of samples per dataset. A summary of the experimental datasets is presented in Table~\ref{table: experimental_data}, with detailed descriptions provided below.

\textbf{SAT-6}\cite{basu2015deepsat} consists of over $0.4$ million images with four bands covering six land cover classes, each sized at $28\times28$ pixels. In our experiments, we randomly selected $4200$ samples from the training data to create the training dataset and $1200$ samples for the validation dataset. To construct the test dataset, $200$ samples per class were randomly selected from the test data. Each image was resized to $32\times32$ by padding with zeros. Consequently, we prepared three balanced datasets containing $4200$, $1200$, and $1200$ samples for training, validation, and testing, respectively.

\textbf{LCZ42}\cite{zhu2019so2sat} originally comprises approximately half a million Sentinel-1 and Sentinel-2 images from $42$ cities, each labeled with $17$ LCZ labels and sized at $32\times32$ pixels. For our experiments, we focused on the Sentinel-2 images containing four bands~(Red, Green, Blue, and Near-infrared) from three German cities: Berlin, Munich, and Cologne. Additionally, we regrouped the LCZ labels into five semantic classes based on the land cover classification scheme~\cite{qiu2019local}. For each class, $2000$ randomly selected images were divided into three subsets: $60\%$ for training, $20\%$ for validation, and $20\%$ for testing, ensuring no overlap among the sets.

\textbf{EuroSAT}\cite{helber2019eurosat} is based on Sentinel-2 imagery covering $13$ bands. It contains a total of $27000$ samples, each with a resolution of $64\times64$ pixels, distributed from ten distinct land cover classes. In the experiments, we focused on five classes~(‘river,’ ‘forest,’ ‘residential,’ ‘sea/lake,’ and ‘industrial’) and randomly selected $5000$ samples with three bands~(Red, Green, and Blue). These samples were then split into training, validation, and test datasets with the ratio $60\%:20\%:20\%$.

\textbf{PatternNet}\cite{zhou2018patternnet} contains high-resolution imagery from $38$ different classes, with $800$ images per class. Each image has three bands and is sized at $256\times256$ pixels. For this dataset, we selected ‘freeway,’ ‘overpass,’ ‘intersection,’ and ‘closed road’ as the target classes. A total of $3200$ images were randomly chosen with a balanced distribution across the classes and split into training, validation, and test datasets in a ratio of $60\%:20\%:20\%$.

\subsection{Experimental Models}
In the experiments, due to the limitations of available quantum resources, our model utilized $12$ qubits: $9$ qubits for encoding the processed image, with a size of $8\times8\times9$, and $3$ qubits for the quantum convolution operations that extracted a feature vector for the final classifier. Specifically, for encoding, larger input images require an increase in the patch size to ensure the processed image maintains a size of $8\times8\times9$. Table~\ref{table: experimental_data} details the patch size for each experimental dataset. For the quantum convolution operations, our model employed two convolution blocks. In these blocks, one qubit represented the indices of the kernels and corresponding feature maps, while the other two qubits encoded the values of the feature maps. After quantum measurement, a feature vector with $64$ values was obtained, which served as the input for the final classifier. The classifier consisted of a single dense layer responsible for the final classification.

As for the competitors, we mainly adopted three different models to investigate the properties of our approach. Specifically, the first model, MLTCNN, is the classical counterpart of our proposed model and also incorporates multitask learning for classification purposes. While its structure is identical to our proposed model, it replaces the quantum convolutional operations with classical convolutional operations to extract the feature vector for final classification. The second model, SEQNN\cite{fan2025hybrid}, is designed to analyze images while minimizing the use of quantum resources. Similar to our model, it splits the input image into patches but adopts a multilayer perceptron module to generate corresponding superpixels for quantum encoding and feature extraction. Thus, unlike our model, SEQNN is based on single-task learning. In our experiments, this model also utilized $12$ qubits, with $9$ for data encoding and $3$ for feature extraction, aligning with our model's quantum resource allocation. The third model, CNN, is a widely used architecture for EO data classification that relies on convolutional operations to extract significant features. Since convolutional operations are also a core component of our proposed model, CNN serves as a baseline for comparison. To ensure a fair and valid comparison, all these experimental models were constructed with a similar number of trainable parameters.

\subsection{Experimental Settings}
We trained the aforementioned model for $200$ epochs with a learning rate of $0.01$, using a batch size of $50$. Each model was trained three times, and for each training, the model with the lowest validation loss value was used to validate its performance on the test dataset. Eventually, the average classification accuracy will be compared and discussed. \F{As for the models utilizing quantum computing, since we aim to evaluate their validity for classification, we employed a noiseless simulator provided by the TensorFlow Quantum platform~\cite{broughton2020tensorflow}, which generates analytic results and allows us to verify the validity of these hybrid models without considering the noise effects.}

\begin{table}[t]
\centering
\caption{\F{Model Performance Across Varying Hyperparameter}}
\setlength{\tabcolsep}{2.5pt}
\begin{tabular}{cccccc}
\toprule
$\alpha$ & SAT-6 & LCZ42 & EuroSAT & PatternNet \\ \midrule
1 & $0.954 \pm 0.006$ & $0.935 \pm 0.004$ & $0.948 \pm 0.003$ & $0.961 \pm 0.022$ \\ 
3 & $0.956 \pm 0.005$ & $0.940 \pm 0.002$ & $0.947 \pm 0.002$  & $0.959 \pm 0.002$ \\ 
5 & $0.958 \pm 0.002$ & $0.941 \pm 0.003$ & $0.954 \pm 0.003$ & $0.963 \pm 0.017$ \\ 
7 & $0.950 \pm 0.001$ & $0.936 \pm 0.003$ & $0.952 \pm 0.001$ & $0.969 \pm 0.007$ \\ 
10 & $0.950 \pm 0.011$ & $0.934 \pm 0.008$ & $0.948 \pm 0.008$ & $0.946 \pm 0.007$ \\ \bottomrule
\end{tabular}
\label{table: finetuning}
\end{table}

\begin{table}[t]
\centering
\caption{\F{Comparison of classification performance between the proposed model and others: $M$ = million, $K$ = thousand.}}
\setlength{\tabcolsep}{5pt}
\begin{tabular}{ccr|lc}
\toprule
Data  & Model & All Para. & Quantum Para. & Test Acc. \\ \midrule
\multirow{4}{*}{SAT-6}  & ours   &  $1.9K$ &  $198$  &  $0.958 \pm 0.002$   \\
                        & MLTCNN &  $1.9K$ & \textbackslash{} &  $0.935 \pm 0.007$   \\
                        & SEQNN\cite{fan2025hybrid}  &  $2.0K$ & $144$ &  $0.942 \pm 0.004$   \\
                        & CNN    &  $1.9K$ & \textbackslash{} &  $0.937 \pm 0.007$   \\ \midrule
\multirow{4}{*}{LCZ42}  & ours   &  $1.8K$ & $198$  &  $0.941 \pm 0.003$   \\
                        & MLTCNN &  $1.8K$ & \textbackslash{} &  $0.910 \pm 0.006$   \\
                        & SEQNN\cite{fan2025hybrid}  &  $2.0K$ & $144$ &  $0.915 \pm 0.009$   \\
                        & CNN    &  $1.8K$ & \textbackslash{} &  $0.919 \pm 0.007$   \\ \midrule
\multirow{4}{*}{EuroSAT}    & ours   &  $3.6K$ & $198$ &  $0.954 \pm 0.003$   \\
                            & MLTCNN &  $3.6K$ & \textbackslash{} &  $0.917 \pm 0.024$   \\
                            & SEQNN\cite{fan2025hybrid}  &  $13.3K$ & $144$ &  $0.890 \pm 0.010$   \\
                            & CNN    &  $3.7K$ & \textbackslash{} &  $0.903 \pm 0.035$   \\  \midrule
\multirow{4}{*}{PatternNet} & ours   &  $10.6K$ & $198$ &  $0.969 \pm 0.007$   \\
                            & MLTCNN &  $10.6K$ & \textbackslash{} &  $0.921 \pm 0.005$   \\
                            & SEQNN\cite{fan2025hybrid}  &  $1.7M$ & $144$ &  $0.777 \pm 0.013$   \\
                            & CNN    &  $10.9K$ & \textbackslash{} &  $0.921 \pm 0.006$   \\  \bottomrule
\end{tabular}
\label{table: performance}
\end{table}

\begin{table*}[t]
\centering
\caption{\F{Model Performance Across Varying Training Sample Sizes}}
\setlength{\tabcolsep}{10.5pt}
\begin{tabular}{cccccccc}
\toprule
 \begin{tabular}[c]{@{}c@{}} Dataset \\ (Available Samples) \end{tabular}   & Model  & \begin{tabular}[c]{@{}c@{}}10\% Training \\ Samples \\ Test Acc. \end{tabular} & \begin{tabular}[c]{@{}c@{}}Test Acc. \\ Difference \end{tabular} & \begin{tabular}[c]{@{}c@{}}50\% Training \\ Samples \\ Test Acc. \end{tabular} & \begin{tabular}[c]{@{}c@{}}Test Acc. \\ Difference \end{tabular} & \begin{tabular}[c]{@{}c@{}}100\% Training \\ Samples  \\ Test Acc.\end{tabular} & \begin{tabular}[c]{@{}c@{}}Test Acc.\\ Difference  \end{tabular} \\ \midrule
\multirow{4}{*}{\begin{tabular}[c]{@{}c@{}} SAT-6  \\ (4200) \end{tabular} } & ours & $0.912 \pm 0.003$ &  \textbackslash{} & $0.945 \pm 0.012$ &  \textbackslash{} &  $0.958 \pm 0.002$ &  \textbackslash{} \\ \cmidrule{2-8}
 & MLTCNN & $0.859 \pm 0.014$ & $0.053 \downarrow$ & $0.919 \pm 0.007$ & $0.026 \downarrow$ &  $0.935 \pm 0.007$ & $0.023 \downarrow$ \\
 & SEQNN\cite{fan2025hybrid} & $0.884 \pm 0.017$ & $0.028 \downarrow$ & $0.908 \pm 0.021$ & $0.037 \downarrow$ &  $0.942 \pm 0.004$ & $0.016 \downarrow$\\
 & CNN & $0.882 \pm 0.010$ & $0.030 \downarrow$ & $0.923 \pm 0.008$ & $0.022 \downarrow$ &  $0.937 \pm 0.007$ & $0.021 \downarrow$ \\ \midrule
\multirow{4}{*}{\begin{tabular}[c]{@{}c@{}} LCZ42  \\ (6000) \end{tabular} } & ours & $0.903 \pm 0.005$ &  \textbackslash{} & $0.932 \pm 0.002$ &  \textbackslash{} &  $0.941 \pm 0.003$ &  \textbackslash{} \\ \cmidrule{2-8}
 & MLTCNN & $0.837 \pm 0.006$ & $0.066 \downarrow$ &  $0.895 \pm 0.006$ & $0.037 \downarrow$ &  $0.910 \pm 0.006$ & $0.031 \downarrow$ \\
 & SEQNN\cite{fan2025hybrid} & $0.861 \pm 0.026$ & $0.042 \downarrow$ &  $0.892 \pm 0.007$ & $0.040 \downarrow$ &  $0.915 \pm 0.009$ & $0.026 \downarrow$ \\
 & CNN & $0.868 \pm 0.005$ & $0.035 \downarrow$ &  $0.907 \pm 0.001$ & $0.025 \downarrow$ &  $0.919 \pm 0.007$ & $0.022 \downarrow$ \\ \midrule
\multirow{4}{*}{\begin{tabular}[c]{@{}c@{}} EuroSAT  \\ (3000) \end{tabular}} & ours  & $0.870 \pm 0.011$ &  \textbackslash{} & $0.947 \pm 0.003$ &  \textbackslash{} &  $0.954 \pm 0.003$ &  \textbackslash{} \\ \cmidrule{2-8}
 & MLTCNN & $0.749 \pm 0.021$ & $0.121 \downarrow$ &  $0.883 \pm 0.007$ & $0.064 \downarrow$ &  $0.917 \pm 0.024$ & $0.037 \downarrow$ \\
 & SEQNN\cite{fan2025hybrid} & $0.736 \pm 0.124$ & $0.134 \downarrow$ &  $0.878 \pm 0.017$ & $0.069 \downarrow$ &  $0.890 \pm 0.010$ & $0.064 \downarrow$ \\
 & CNN &  $0.792 \pm 0.023$ & $0.078 \downarrow$ &  $0.888 \pm 0.016$ & $0.059 \downarrow$ &  $0.903 \pm 0.035$ &  $0.051 \downarrow$\\ \midrule
\multirow{4}{*}{\begin{tabular}[c]{@{}c@{}} PatternNet \\ (1920) \end{tabular} } & ours & $0.868 \pm 0.023$ &  \textbackslash{} & $0.959 \pm 0.001$ &  \textbackslash{} &  $0.969 \pm 0.007$ &  \textbackslash{} \\ \cmidrule{2-8}
 & MLTCNN & $0.591 \pm 0.026$ & $0.277 \downarrow$  &  $0.870 \pm 0.007$ & $0.089 \downarrow$ &  $0.921 \pm 0.005$ & $0.048 \downarrow$\\
 & SEQNN\cite{fan2025hybrid} & $0.699 \pm 0.046$ & $0.169 \downarrow$  &  $0.747 \pm 0.004$ & $0.212 \downarrow$ &  $0.777 \pm 0.013$ & $0.192 \downarrow$ \\
 & CNN & $0.620 \pm 0.121$ & $0.248 \downarrow$ &  $0.914 \pm 0.023$ & $0.045 \downarrow$ &  $0.921 \pm 0.006$ & $0.048 \downarrow$\\ \bottomrule
\end{tabular}
\label{table: training_Samples}
\end{table*}

\begin{table*}[t]
\centering
\caption{Model Performance with imbalanced training data}
\setlength{\tabcolsep}{19.5pt}
\begin{tabular}{cccccc}
\toprule
Model & Setting & SAT-6 & LCZ42 & EuroSAT & PatternNet \\ \midrule
\multirow{3}{*}{Ours} & Balanced  & $0.958 \pm 0.002$ & $0.941 \pm 0.003$ & $0.954 \pm 0.003$ & $0.969 \pm 0.007$ \\
 & Imbalanced & $0.930 \pm 0.015$ & $0.904 \pm 0.016$ & $0.918 \pm 0.018$ & $0.890 \pm 0.044$ \\
 & Minor Class F1 & $0.873 \pm 0.076$ & $0.835 \pm 0.099$ & $0.851 \pm 0.049$ & $0.762 \pm 0.163$ \\ \midrule
\multirow{3}{*}{MLTCNN} & Balanced & $0.935 \pm 0.007$ & $0.910 \pm 0.006$ & $0.917 \pm 0.024$ & $0.921 \pm 0.005$ \\
 & Imbalanced & $0.895 \pm 0.023$ & $0.857 \pm 0.027$ & $0.834 \pm 0.033$ & $0.810 \pm 0.037$ \\
 & Minor Class F1 & $0.776 \pm 0.138$ & $0.715 \pm 0.177$ & $0.665 \pm 0.117$ & $0.618 \pm 0.167$ \\ \midrule
\multirow{3}{*}{SEQNN\cite{fan2025hybrid}} & Balanced & $0.942 \pm 0.004$ & $0.915 \pm 0.009$ & $0.890 \pm 0.010$ & $0.777 \pm 0.013$ \\
 & Imbalanced & $0.907 \pm 0.030$ & $0.864 \pm 0.022$ & $0.795 \pm 0.048$ & $0.640 \pm 0.143$ \\
 & Minor Class F1 & $0.804 \pm 0.153$ & $0.754 \pm 0.153$ & $0.438 \pm 0.258$ & $0.304 \pm 0.370$ \\ \midrule
\multirow{3}{*}{CNN} & Balanced & $0.937 \pm 0.007$ & $0.919 \pm 0.007$ & $0.903 \pm 0.035$ & $0.921 \pm 0.006$ \\
 & Imbalanced & $0.908 \pm 0.028$ & $0.880 \pm 0.029$ & $0.856 \pm 0.033$ & $0.855 \pm 0.036$ \\
 & Minor Class F1 & $0.799 \pm 0.136$ & $0.773 \pm 0.163$ & $0.672 \pm 0.125$ & $0.700 \pm 0.147$ \\ \bottomrule
\end{tabular}
\label{table: training_imbalance}
\end{table*}

\subsection{Hyperparameter Studies}\label{Hyperparameter}

As described in Section \ref{methodology}, a hyperparameter $\alpha$ is introduced to balance the training between the main task and the auxiliary task. To assess its impact on the classification performance of our model, we conducted experiments with various $\alpha$ values. Their results are summarized in Table~\ref{table: finetuning}. 

As indicated in the table, we experimented with five different $\alpha$ values for each experimental dataset. The results indicate that the predefined $\alpha$ value does influence the classification performance, but the differences are relatively minor, with a maximum performance difference of $0.023$ for PatternNet, $0.008$ for SAT-6, and $0.007$ for LCZ42 and EuroSAT. This suggests that our model demonstrates robustness with respect to this hyperparameter.

\subsection{Model Performance Studies}
To evaluate the effectiveness of our approach to classifying EO data, we compared our model against three other models in terms of test accuracy. Table~\ref{table: performance} presents the experimental results. As shown in the table, our model, despite having fewer trainable parameters, outperforms competitors with a higher parameter count. It highlights our model's ability to effectively extract critical features from EO data for classification.

\subsection{Generalizability Studies}
Regarding the generalizability of our model, we conducted evaluations under two challenging scenarios commonly encountered in EO data analysis: limited training samples and imbalanced training samples. A highly generalizable model is expected to perform well on unseen data in such scenarios.

\subsubsection{The number of training samples} Labeled EO data are often scarce due to the high costs of expert annotation. A model with high generalizability effectively utilizes limited training samples to achieve robust performance on unseen data, which is particularly valuable for EO applications. 

To evaluate our model's generalizability under this scenario, we conducted three experiments for each dataset, using $10\%$, $50\%$, and $100\%$ of the training samples to train the models while keeping the test datasets unchanged for evaluation. The results, presented in Table~\ref{table: training_Samples}, show that the performance of all models improves as the number of training samples increases—a trend widely reported in other studies. Notably, our model consistently outperforms its competitors, with its advantages becoming particularly evident when training samples are limited.

\begin{table*}[t]
\centering
\caption{Ablation Study Results for our model}
\setlength{\tabcolsep}{9.5pt}
\begin{tabular}{ccccccc}
\toprule
Setting & \begin{tabular}[c]{@{}c@{}}Reconstruction\\ Branch\end{tabular} & \begin{tabular}[c]{@{}c@{}}Location Weight \\ Module\end{tabular} & SAT-6 & LCZ42 & EuroSAT & PatternNet \\ \midrule
\multirow{4}{*}{10\% Training Samples} & \checkmark &  & $0.910 \pm 0.001$ & $0.891 \pm 0.003$ & $0.868 \pm 0.011$ & $0.829 \pm 0.048$ \\ \cmidrule{2-7} 
 &  & \checkmark & $0.891 \pm 0.001$ & $0.899 \pm 0.012$ & $0.862 \pm 0.017$ & $0.791 \pm 0.076$ \\ \cmidrule{2-7} 
 & \checkmark & \checkmark & $0.912 \pm 0.003$ & $0.903 \pm 0.005$ & $0.870 \pm 0.011$ & $0.868 \pm 0.023$ \\ \midrule
\multirow{4}{*}{50\% Training Samples} & \checkmark &  & $0.938 \pm 0.013$ & $0.925 \pm 0.006$ & $0.936 \pm 0.009$ & $0.916 \pm 0.024$ \\ \cmidrule{2-7} 
 &  & \checkmark & $0.939 \pm 0.003$ & $0.929 \pm 0.007$ & $0.936 \pm 0.012$ & $0.942 \pm 0.029$ \\ \cmidrule{2-7} 
 & \checkmark & \checkmark & $0.945 \pm 0.012$ & $0.932 \pm 0.002$ & $0.947 \pm 0.003$ & $0.959 \pm 0.001$ \\ \midrule
\multirow{4}{*}{100\% Training Samples} & \checkmark &  & $0.951 \pm 0.010$ & $0.933 \pm 0.003$ & $0.941 \pm 0.009$ & $0.932 \pm 0.053$ \\ \cmidrule{2-7} 
 &  & \checkmark & $0.946 \pm 0.010$ & $0.936 \pm 0.001$ & $0.941 \pm 0.007$ & $0.976 \pm 0.005$ \\ \cmidrule{2-7} 
 & \checkmark & \checkmark & $0.958 \pm 0.002$ & $0.941 \pm 0.003$ & $0.954 \pm 0.003$ & $0.969 \pm 0.007$ \\ \bottomrule
\end{tabular}
\label{table: ablation}
\end{table*}

\subsubsection{The imbalanced training samples} Imbalanced datasets are frequently encountered in EO tasks, such as land cover classification, where real-world class distributions are often uneven. Training on such datasets can introduce biases, negatively affecting the model's performance on underrepresented classes. A highly generalizable model can mitigate these biases, leading to improved performance

In the case of imbalanced training data, we conducted multiple experiments for each dataset. In each experiment, one category was designated as the minority class, using only $10\%$ of its training samples, while the remaining categories used all available samples. For evaluation, all the images from the test dataset were utilized. Table~\ref{table: training_imbalance} shows the experimental results, comparing the test accuracy of the model trained on the imbalanced dataset with that of the model trained on a balanced dataset using all available samples. Additionally, we compared the averaged F1 scores of the minority classes in our evaluation.

As shown in the table, an imbalanced training dataset negatively impacts the overall classification performance of the trained model, as evidenced by the significant difference in test accuracy between models trained on balanced datasets and those trained on imbalanced datasets. Compared to other models, our proposed model achieves higher average overall classification accuracy and higher average F1 scores for the minority classes. This demonstrates its ability to extract meaningful features from the minority classes, further highlighting its generalizability advantages in this context.

\subsection{Ablation Studies}
To assess the contributions of the auxiliary task and the proposed location weight module to the classification performance of our model, we conducted ablation studies by separately removing the image reconstruction component and the location weight module in the quantum circuit for feature extraction. 

Table~\ref{table: ablation} summarizes the results of our studies with varying numbers of training samples. As shown, the classification performance of our model decreases when either the reconstruction branch or the location weight module is omitted, highlighting the importance of these components in achieving optimal performance. 

Notably, when the number of training samples is very limited, the model with the reconstruction branch generally achieves better classification performance compared to the model with only the location weight module. Conversely, as the number of training samples increases, the model with the location weight module outperforms the one with solely the reconstruction branch. This observation suggests that the reconstruction branch is more effective in enhancing classification performance when training data are scarce, whereas the location weight module becomes increasingly beneficial as the amount of training data grows.

\section{Discussion}\label{discussion}

As the experimental results suggest, our model demonstrates superior generalizability compared to the competitors. In this section, we aim to examine the underlying factors contributing to this advantage through a comparative analysis of our model and MLTCNN, the classical counterpart of our model. Specifically, as illustrated in Fig.~\ref{fig: overall}, the classification task involves two classical outputs in the pipeline: the processed image and the extracted feature vector. While the MLTCNN model also produces these two outputs, it relies solely on classical computing to derive the feature vector from the processed image. Thus, these outputs are identical for both models.

In this study, we first evaluated the effectiveness of the extracted feature vector for classification. Subsequently, we analyzed the processed image in combination with the feature vector to assess the contribution of the quantum component to the overall performance.

\begin{figure*}[htbp!]
\centering
\makebox[\textwidth]{\makebox[\textwidth]{%
\begin{minipage}[b]{.255\linewidth}
\centering
\includegraphics[width=\linewidth]{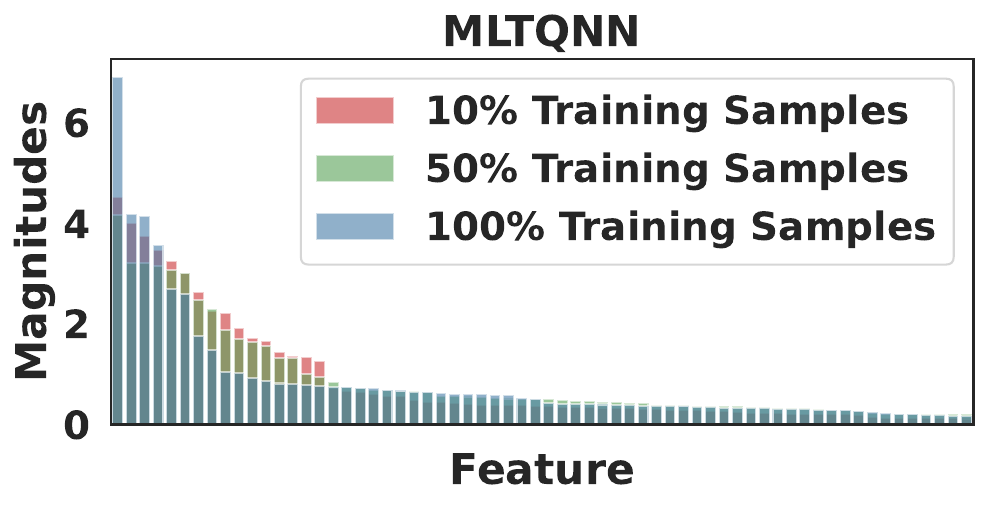}
\subfloat[SAT-6]{%
\includegraphics[width=\linewidth]{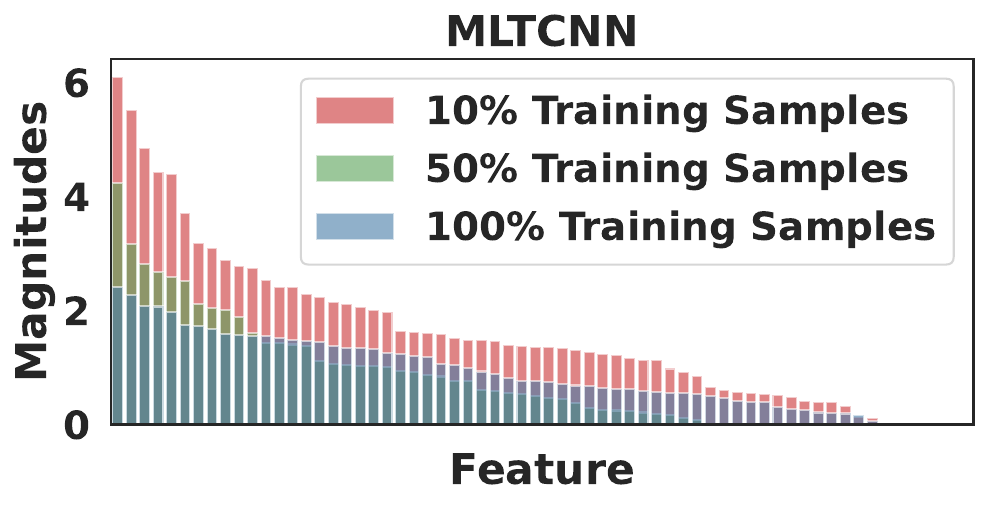}}
\end{minipage}
\begin{minipage}[b]{.255\linewidth}
\centering
\includegraphics[width=\linewidth]{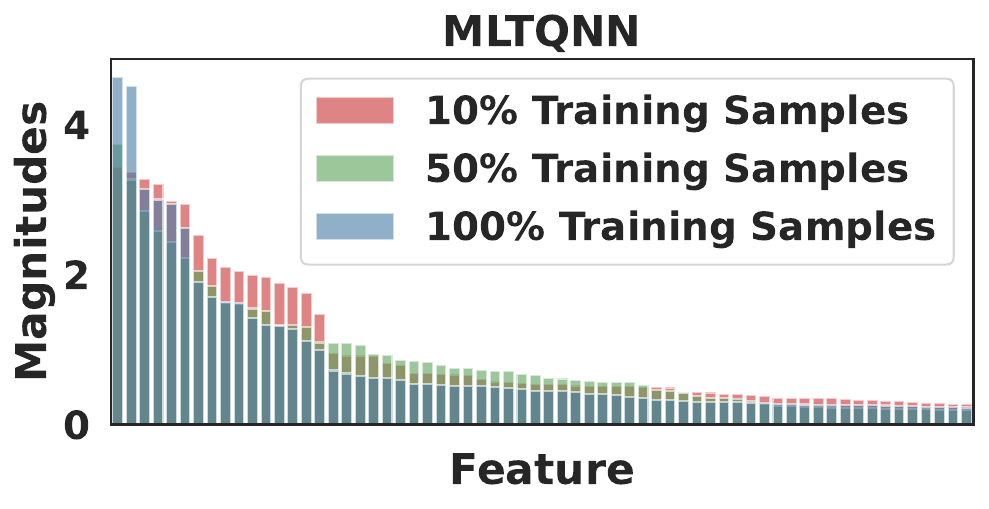}
\subfloat[LCZ42]{%
\includegraphics[width=\linewidth]{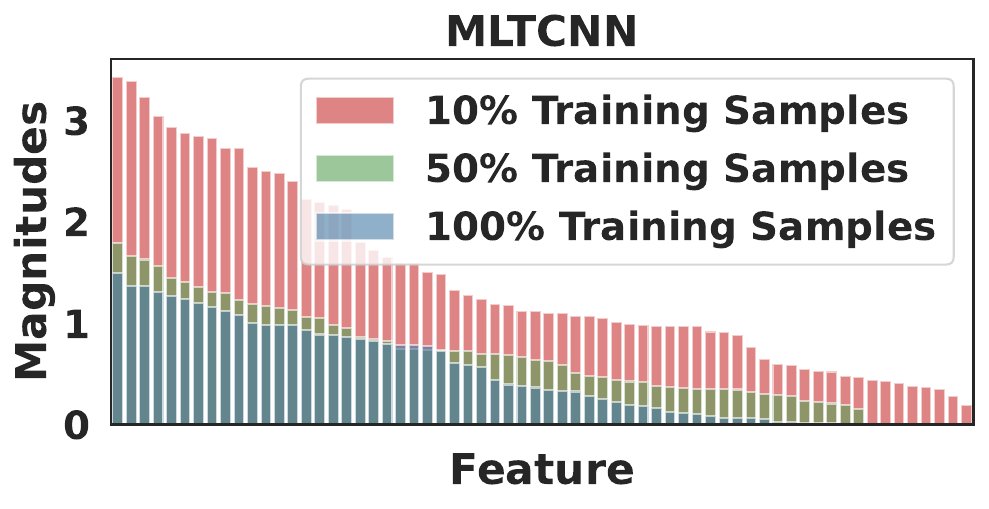}}
\end{minipage}
\begin{minipage}[b]{.255\linewidth}
\centering
\includegraphics[width=\linewidth]{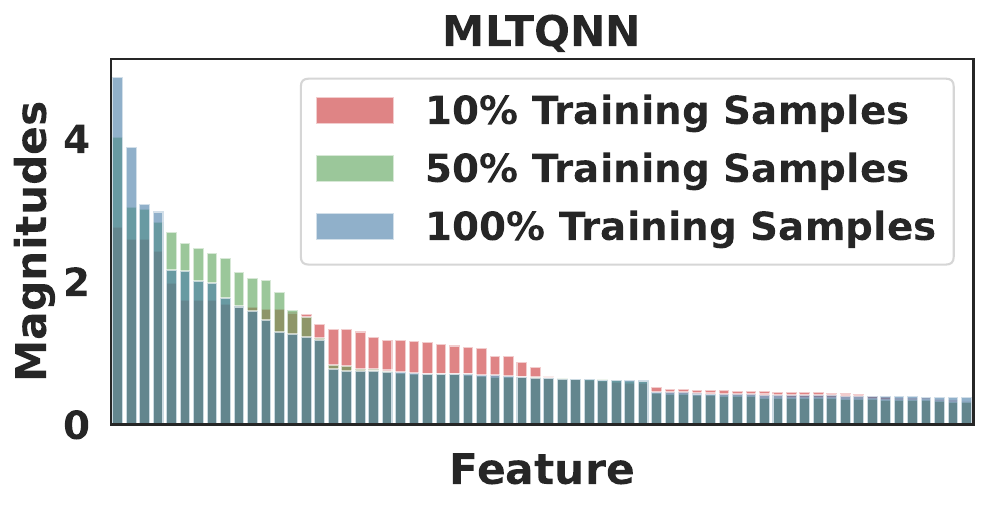}
\subfloat[EuroSAT]{%
\includegraphics[width=\linewidth]{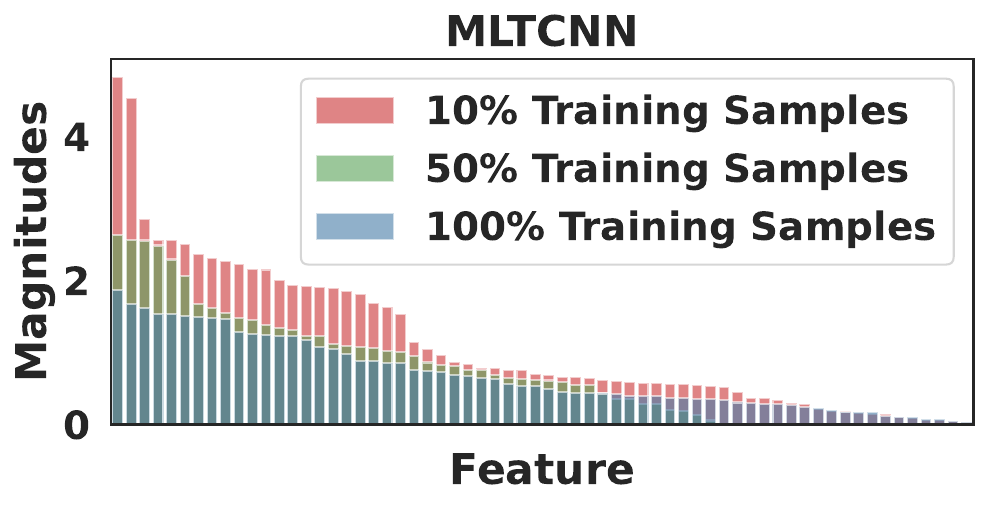}}
\end{minipage}
\begin{minipage}[b]{.255\linewidth}
\centering
\includegraphics[width=\linewidth]{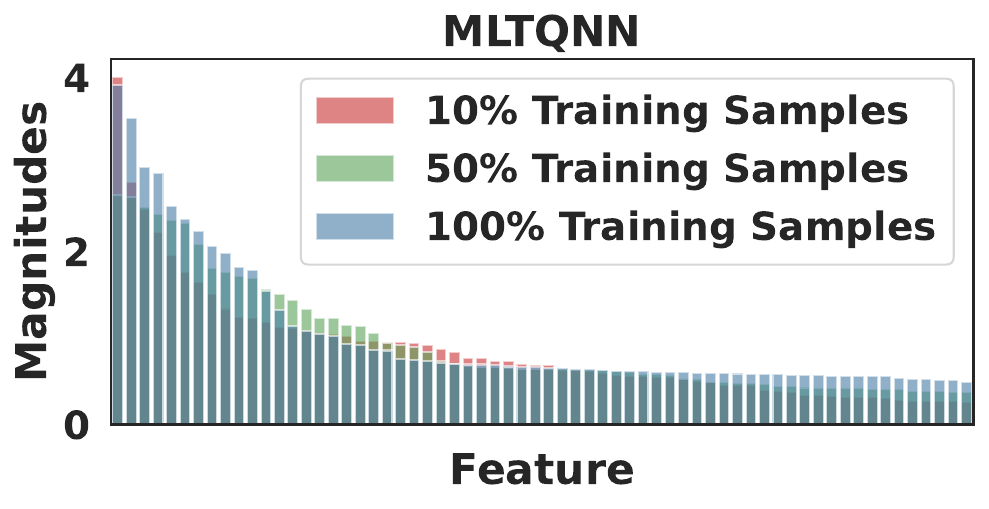}
\subfloat[PatternNet]{%
\includegraphics[width=\linewidth]{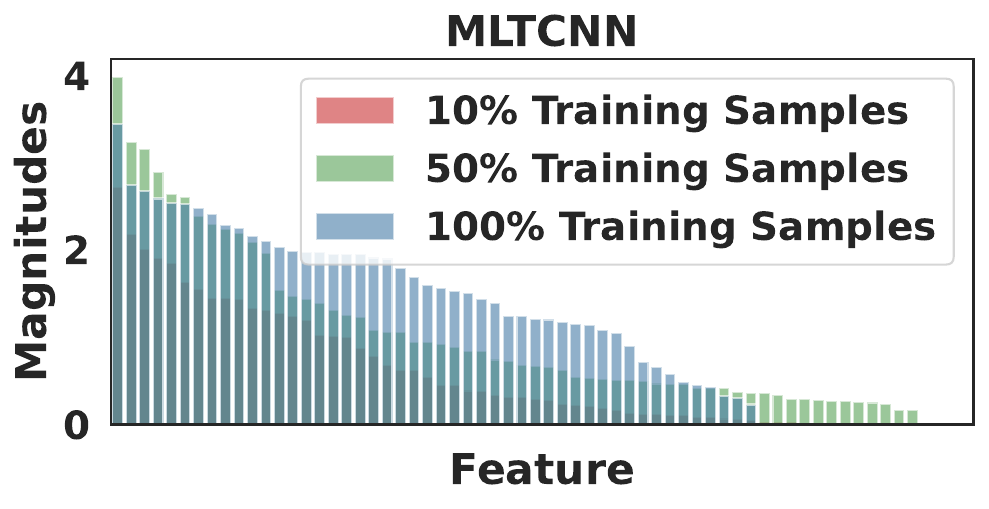}}
\end{minipage}}}    
\caption{Ranked magnitudes of the $64$ extracted features from the trained MLTQNN and MLTCNN models: (a) SAT-6, (b) LCZ42, (c) EuroSAT, (d) PatternNet. Each feature is sorted by its magnitude in descending order, and its rank is shown on the x-axis. The y-axis displays the magnitude value}
\label{fig:feature_magnitude} 
\end{figure*}

\begin{figure*}[htbp!]
\centering
\makebox[\textwidth]{\makebox[\textwidth]{%
\begin{minipage}[b]{.26\linewidth}
\centering
\includegraphics[width=\linewidth]{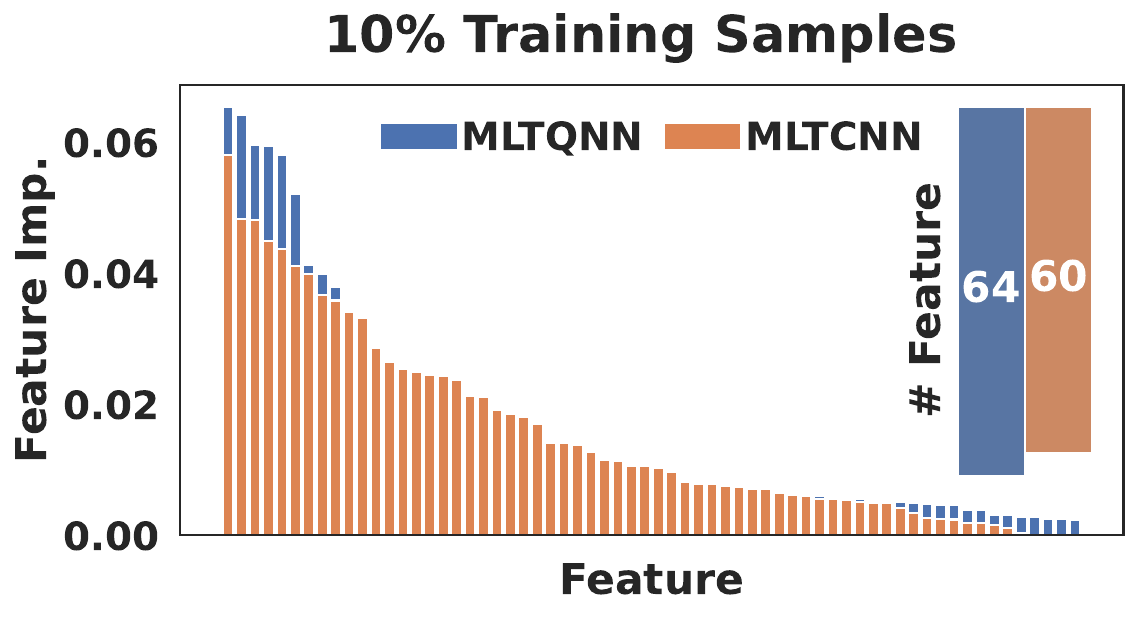}
\includegraphics[width=\linewidth]{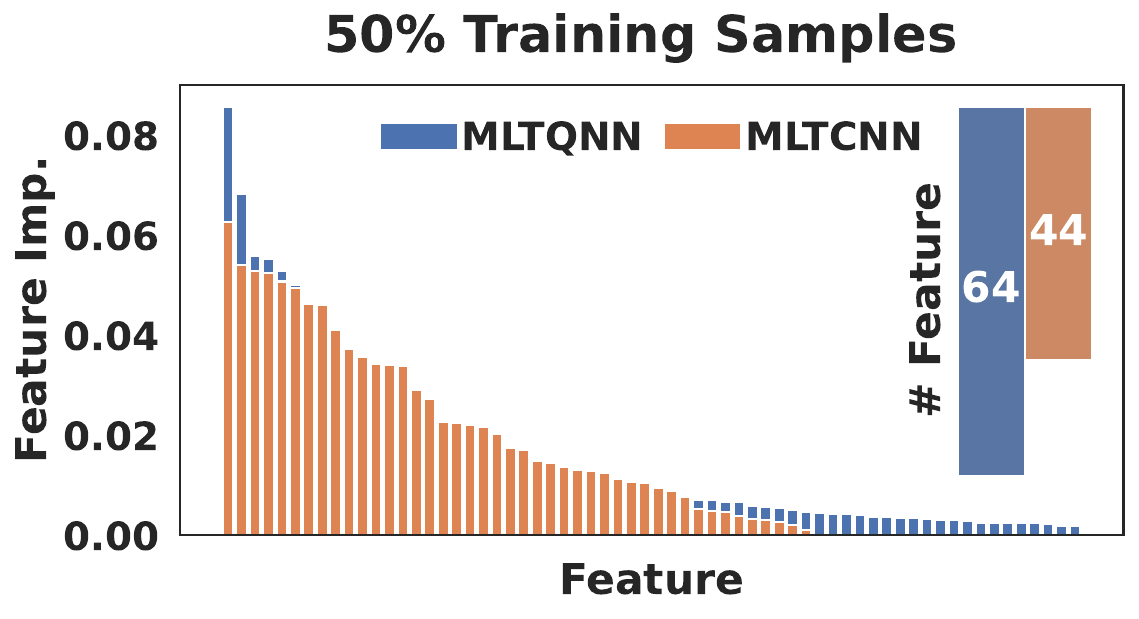}
\subfloat[SAT-6]{%
\includegraphics[width=\linewidth]{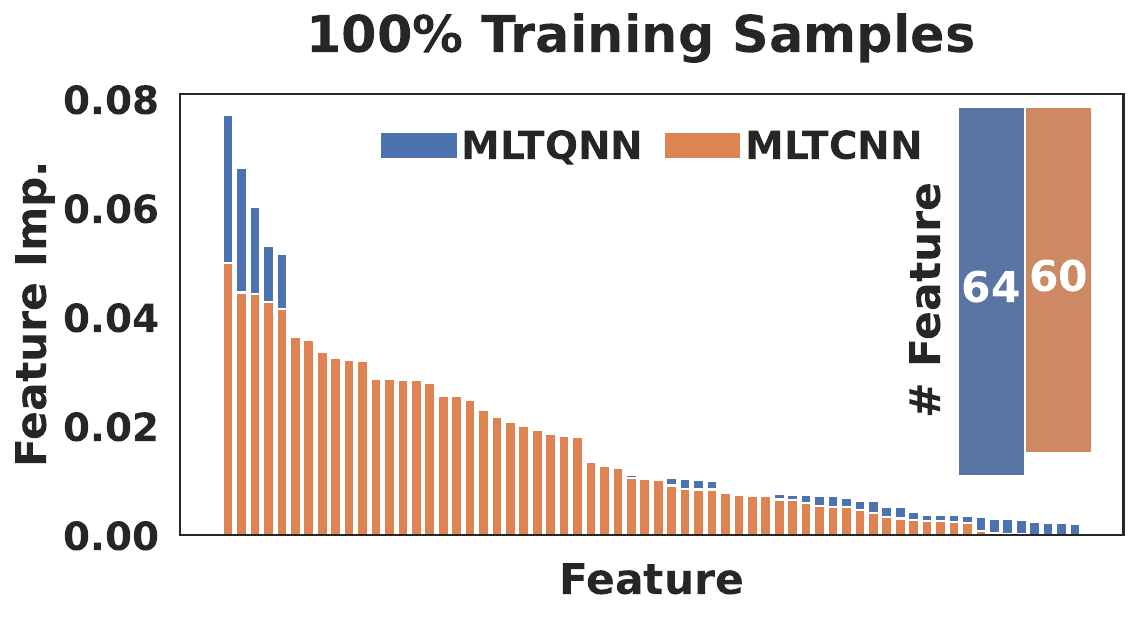}}
\end{minipage}
\begin{minipage}[b]{.26\linewidth}
\centering
\includegraphics[width=\linewidth]{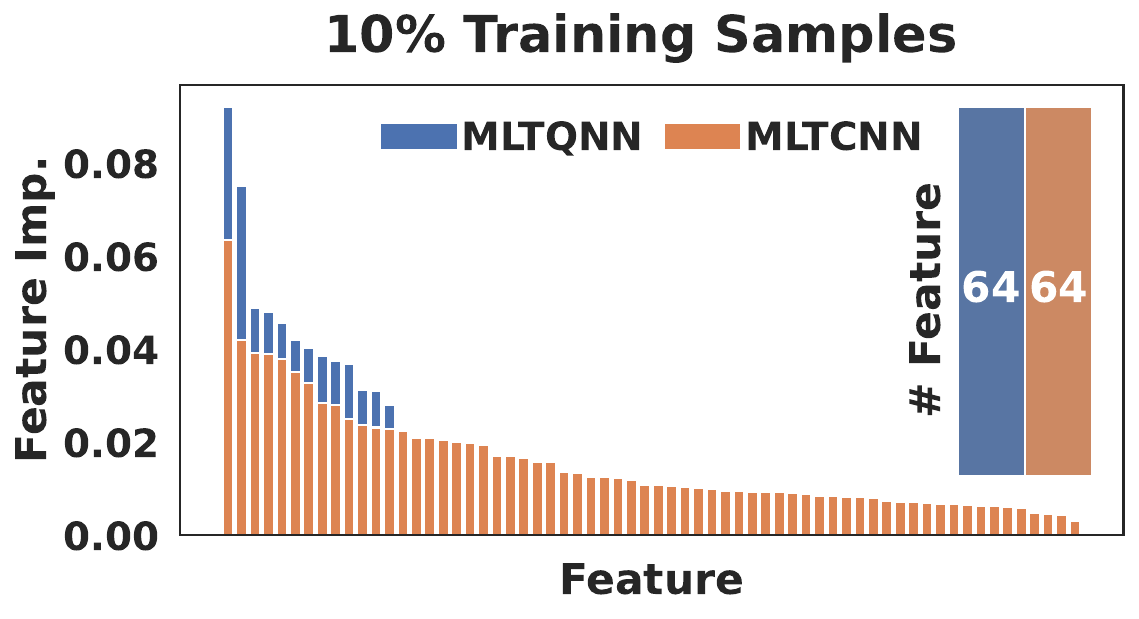}
\includegraphics[width=\linewidth]{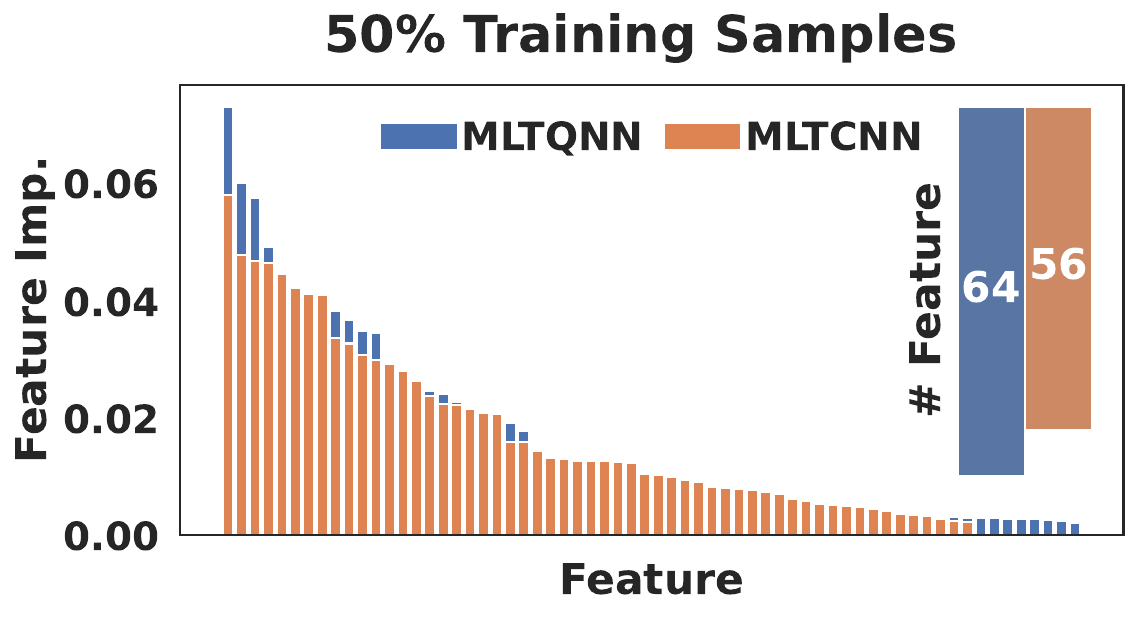}
\subfloat[LCZ42]{%
\includegraphics[width=\linewidth]{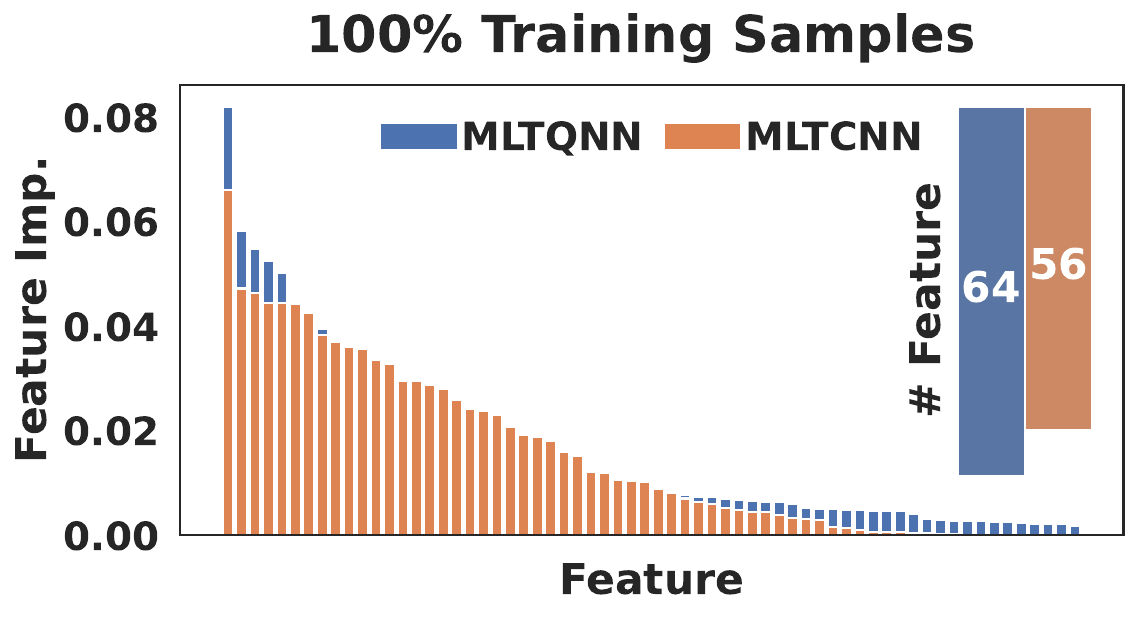}}
\end{minipage}
\begin{minipage}[b]{.26\linewidth}
\centering
\includegraphics[width=\linewidth]{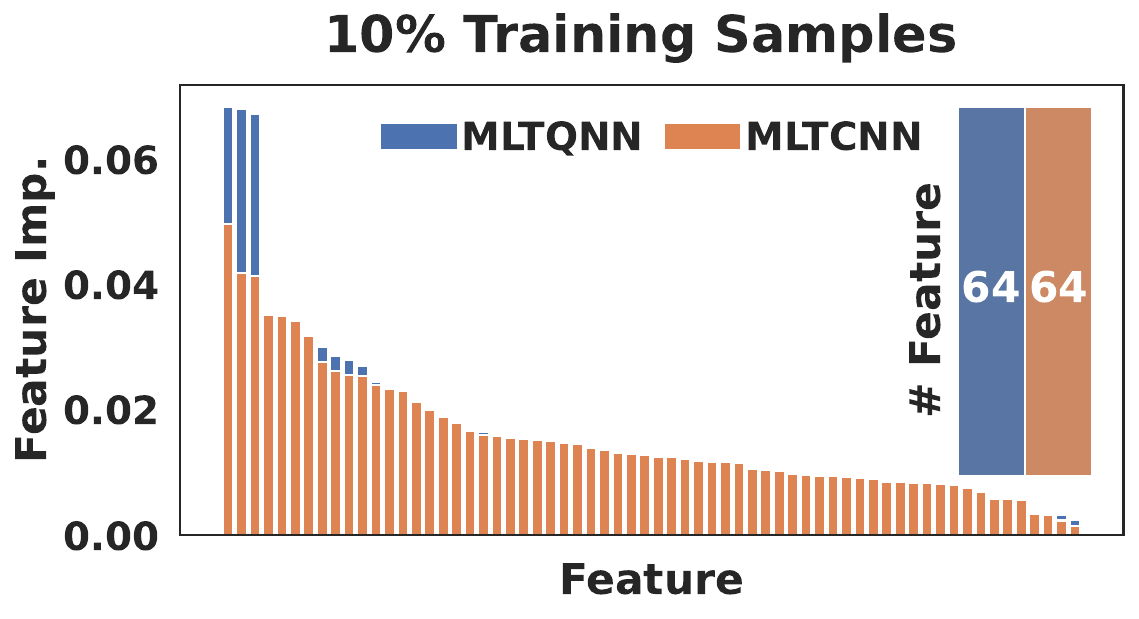}
\includegraphics[width=\linewidth]{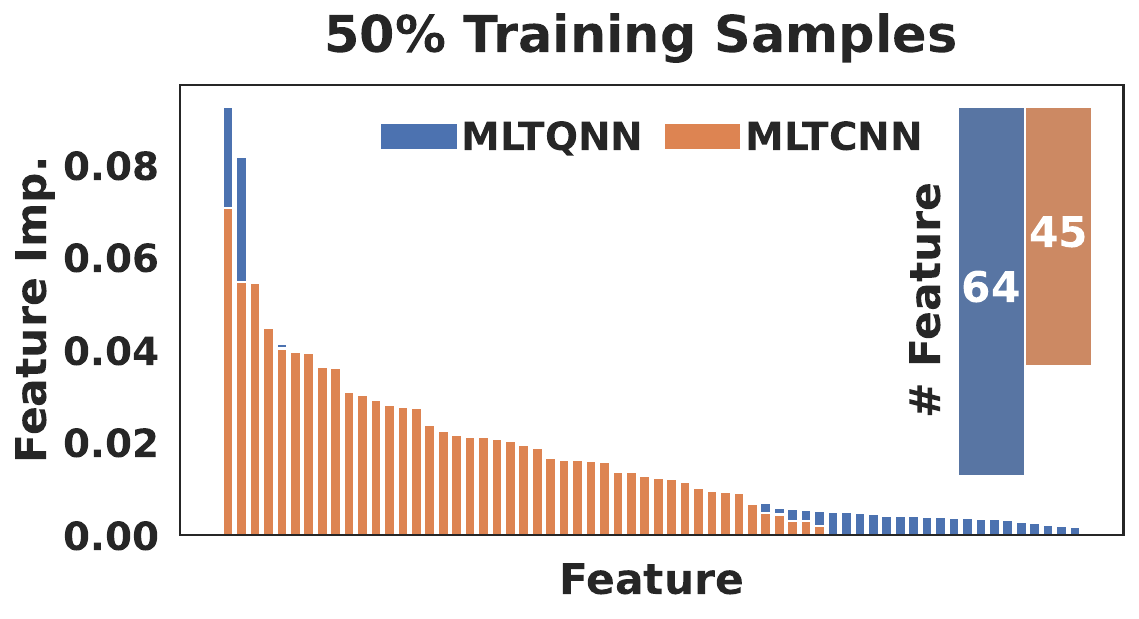}
\subfloat[EuroSAT]{%
\includegraphics[width=\linewidth]{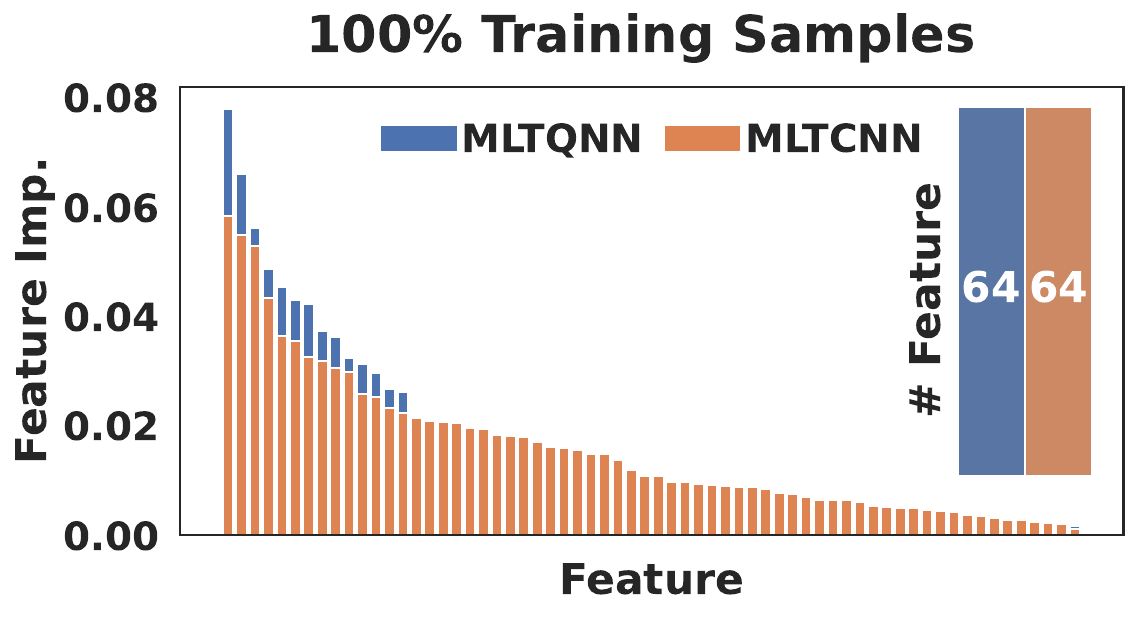}}
\end{minipage}
\begin{minipage}[b]{.26\linewidth}
\centering
\includegraphics[width=\linewidth]{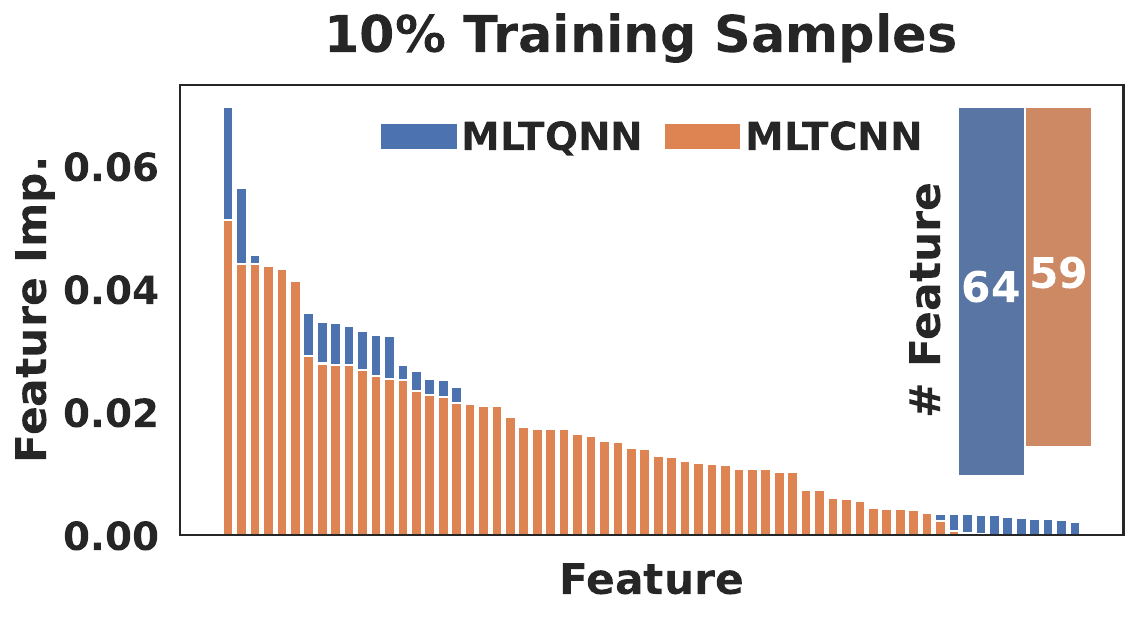}
\includegraphics[width=\linewidth]{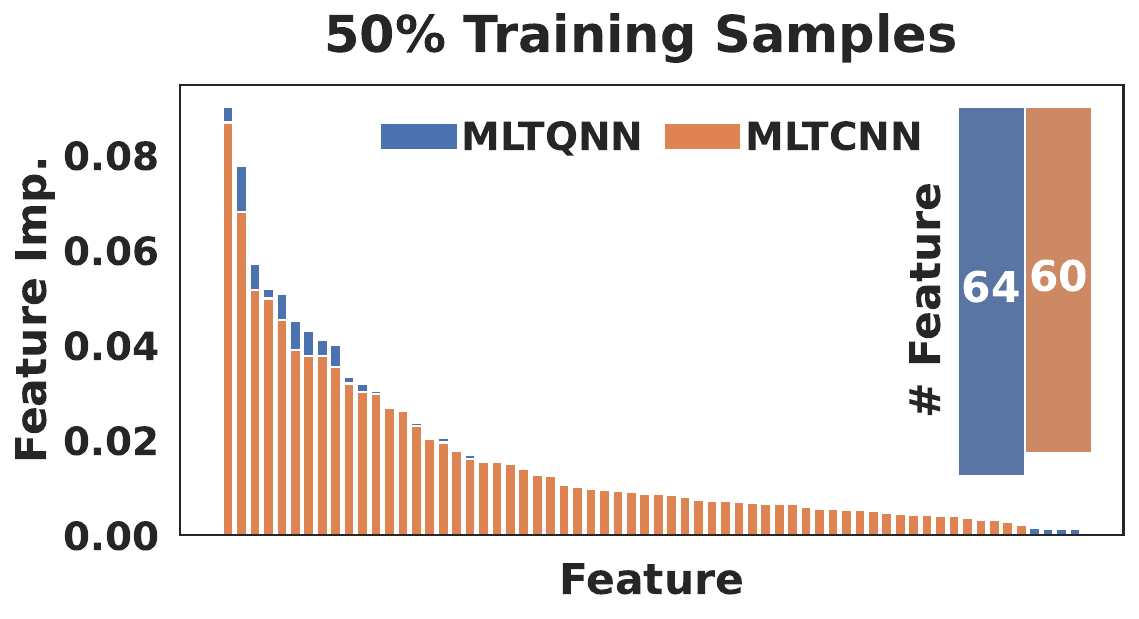}
\subfloat[PatternNet]{%
\includegraphics[width=\linewidth]{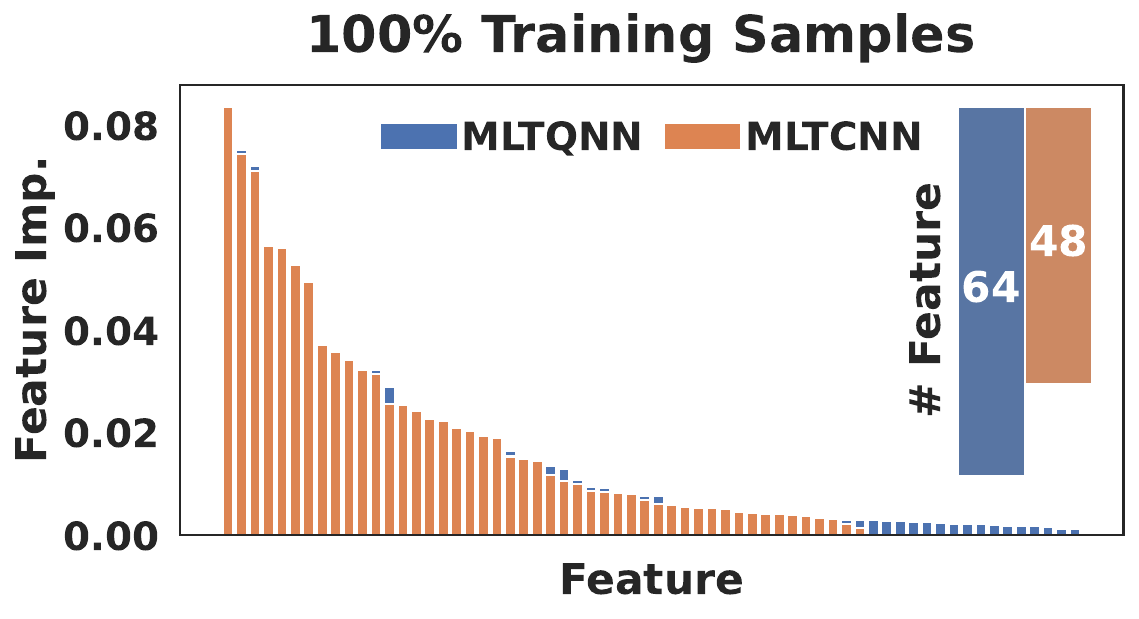}}
\end{minipage}}}    
\caption{Ranked importance values of the $64$ extracted features from the trained MLTQNN \& MLTCNN models and the number of used features (features with importance greater than $0$): (a) SAT-6, (b) LCZ42, (c) EuroSAT, (d) PatternNet. Each feature is assigned an index based on its ranked importance, and the x-axis reflects this index in descending order of importance. The y-axis indicates the corresponding importance value for a visual comparison}
\label{fig:feature_importance} 
\end{figure*}
\begin{table*}[t]
\centering
\setlength{\tabcolsep}{9.5pt}
\caption{IG Values Metrics Across varying training sample sizes}
\begin{tabular}{ccccccccccc}
\toprule
\multirow{3}{*}{Dataset} & \multirow{3}{*}{Model} & \multicolumn{3}{c}{10\% Training Sample} & \multicolumn{3}{c}{50\% Training Sample} & \multicolumn{3}{c}{100\% Training Sample} \\ \cmidrule{3-11} 
 &  & Count   & IG Value &  IG Diff. & Count & IG Value & IG Diff. & Count & IG Value & IG Diff. \\ \midrule
\multirow{2}{*}{SAT-6} & ours & $25.084$ & $2.780$ & $1.798$ & $22.843$ & $2.596$ & $1.665$ & $23.996$ & $2.735$ & $1.808$\\
                       & MLTCNN & $22.760$ & $2.389$ & $1.478$ & $22.084$ & $2.356$ & $1.508$ & $21.359$ & $2.283$ & $1.475$\\ \midrule
\multirow{2}{*}{LCZ42} & ours & $24.516$ & $2.646$ & $1.703$ & $23.998$ & $2.786$ & $1.834$ & $23.946$ & $2.833$ & $1.883$ \\
                       & MLTCNN & $22.883$ & $2.413$ & $1.533$ & $22.264$ & $2.506$ & $1.622$ & $21.968$ & $2.902$ & $1.942$\\ \midrule
\multirow{2}{*}{EuroSAT} & ours & $25.527$ & $2.693$ & $1.760$ & $25.187$ & $2.886$ & $1.950$ & $26.034$ & $2.748$ & $1.851$ \\
                         & MLTCNN & $22.091$ & $2.684$ & $1.678$ & $22.845$ & $2.762$ & $1.780$ & $24.086$ & $2.578$ & $1.658$\\ \midrule
\multirow{2}{*}{PatternNet} & ours & $24.634$ & $2.721$ & $1.730$ & $26.414$ & $2.556$ & $1.683$ & $26.078$ & $2.604$ & $1.709$\\
                            & MLTCNN & $18.978$ & $2.228$ & $1.338$ & $24.198$ & $2.439$ & $1.575$ & $24.384$ & $2.022$ & $1.286$ \\ \bottomrule
\end{tabular}
\label{table:ig_overall}
\end{table*}

\begin{table*}[t]
\centering
\setlength{\tabcolsep}{8.5pt}
\caption{Adjusted Mutual Information (AMI) scores for K-means clustering using the processed images and feature vectors as inputs}
\begin{tabular}{cccccccc}
\toprule
\multirow{3}{*}{Dataset} & \multirow{3}{*}{Model} & \multicolumn{2}{c}{10\% Training Sample} & \multicolumn{2}{c}{50\% Training Sample} & \multicolumn{2}{c}{100\% Training Sample} \\ \cmidrule{3-8} 
 &  & \begin{tabular}[c]{@{}c@{}} Processed Image \\ ($576$ features)\end{tabular}   & \begin{tabular}[c]{@{}c@{}} Feature Vector  \\ ($64$ features)\end{tabular} & \begin{tabular}[c]{@{}c@{}} Processed Image \\ ($576$ features)\end{tabular} & \begin{tabular}[c]{@{}c@{}} Feature Vector  \\ ($64$ features)\end{tabular} & \begin{tabular}[c]{@{}c@{}} Processed Image \\ ($576$ features)\end{tabular} & \begin{tabular}[c]{@{}c@{}} Feature Vector  \\ ($64$ features)\end{tabular} \\ \midrule
\multirow{2}{*}{SAT-6} & ours & $0.734$ & $0.772$ & $0.738$ & $0.785$ & $0.727$ & $0.780$ \\
 & MLTCNN & $0.675$ & $0.703$ & $0.719$ & $0.776$ & $0.718$ & $0.767$ \\ \midrule
\multirow{2}{*}{LCZ42} & ours & $0.506$ & $0.631$ & $0.557$ & $0.677$ & $0.522$ & $0.644$ \\
 & MLTCNN & $0.501$ & $0.531$ & $0.579$ & $0.638$ & $0.552$ & $0.633$ \\ \midrule
\multirow{2}{*}{EuroSAT} & ours & $0.505$ & $0.668$ & $0.570$ & $0.726$ & $0.607$ & $0.740$ \\
 & MLTCNN & $0.487$ & $0.456$ & $0.488$ & $0.621$ & $0.551$ & $0.640$ \\ \midrule
\multirow{2}{*}{PatternNet} & ours & $0.483$ & $0.552$ & $0.598$ & $0.660$ & $0.528$ & $0.688$ \\
  & MLTCNN & $0.319$ & $0.226$ & $0.484$ & $0.502$ & $0.513$ & $0.526$ \\ \midrule
\end{tabular}
\label{table:ami_Scores}
\end{table*}

\subsection{The Validity Study of the Extracted Feature Vector}

To evaluate the significance of the extracted feature vector, we first used the pre-trained MLTQNN and MLTCNN models, trained with varying numbers of training samples and excluding their final classifiers, to generate learned representations of the test dataset images. Next, we employed three different methods to assess and compare their validity.

\subsubsection{Feature Magnitude} For each dataset, the magnitudes of the $64$ features generated by the model trained with varying numbers of samples were compared.

For the analysis, we ranked $64$ generated features based on their magnitudes. As shown in Fig.~\ref{fig:feature_magnitude}, for each dataset, our models trained with $10\%$, $50\%$, and $100\%$ of the samples generated feature vectors with more consistent magnitudes than MLTCNN. This indicates the robustness of our model with respect to the number of training samples and its effect on classification performance. 

\subsubsection{Feature Importance} The importance of the features in the extracted feature vector was assessed using a random forest classifier \cite{breiman2001random}. It was calculated based on the total reduction in Gini impurity contributed by each feature, with higher values indicating greater significance for classification.

Fig.~\ref{fig:feature_importance} illustrates the comparison between our model and its classical counterpart (MLTCNN) in terms of the features ranked by importance value and the number of features with importance values greater than zero. As shown, across all experimental datasets and settings, every feature in the feature vector generated by our model has an importance value greater than zero, indicating that all features contribute to the classification tasks. In contrast, the classical counterpart (MLTCNN) produces fewer features with non-zero importance values. Note that as the sample size increases, the feature importance in MLTCNN does not exhibit a consistent trend. This is likely due to the hyperparameters being fine-tuned individually for each setting, and the model's higher sensitivity to these values compared to MLTQNN. Moreover, the top-ranked features in our model exhibit higher importance values than those in MLTCNN. These findings highlight that our model generates more meaningful and significant representations of input images for classification tasks.

\subsubsection{Guided Integrated Gradient} This technique, introduced in \cite{kapishnikov2021guided}, attributes a model's predictions to its input features. To mitigate the impact of baseline selection, we adopted multiple baselines of size $64$. For each feature in the feature vector of a given image, we calculated multiple IG values and used their average to represent this feature's significance level with respect to the classifier's prediction. Additionally, we computed the average IG values for each feature across all possible categories rather than limiting the analysis to the ground truth or the predicted category. Consequently, we analyzed the relationship between the IG values corresponding to the ground truth and those associated with other categories to gain deeper insights into the validity of the extracted feature vectors.

To be more specific, for each feature in the feature vector of an input image, we calculated multiple IG values, each corresponding to a possible category. If the highest IG value was associated with the ground truth label, we identified this feature as a valid feature and calculated the difference between this IG value and the second-highest IG value. Accordingly, for each input image, we computed three metrics: the number of valid features, the sum of their IG values, and the sum of the IG value differences. For each dataset, we averaged these metrics across the test samples to enable comparative analysis. Table~\ref{table:ig_overall} summarizes the numerical results of these focused metrics for each dataset under different training settings.

As indicated in the table, the feature vectors generated by our model contain a higher number of valid features compared to those extracted by the MLTCNN. Furthermore, the IG values of these valid features, along with their corresponding IG value differences, are consistently larger for our model. This highlights our approach's ability to produce more significant and distinguishable feature representations for the final classifier.

The findings from the above three approaches are consistent, further supporting the representational advantage of the feature vector generated by our approach over MLTCNN. Additionally, the comparison of the feature vectors generated by the model trained with different numbers of training samples indicates that our model can produce more consistent and valid features, even with a limited number of training samples, underscoring its generalizability advantages.

\subsection{The Validity Study of the Quantum Component}

To evaluate the contribution of the quantum component in our model to classification performance, we assessed and compared the validity of its input (the processed image) and its output (the extracted feature vector).

Specifically, for each dataset, we first used the pre-trained MLTQNN and MLTCNN models, optimized with $10\%$, $50\%$, and $100\%$ of the training samples, to generate the processed images and the extracted feature vectors for the samples in the test dataset. Next, we applied the unsupervised k-means algorithm to cluster the flattened processed images and feature vectors. Using the ground truth for the test dataset, we computed the adjusted mutual information (AMI) score to quantify the agreement between the ground truth and the clustering assignments. An AMI score close to one suggests strong agreement, while a score close to zero indicates significant independence. 

Table~\ref{table:ami_Scores} summarizes the results. As shown in the table, the clusters formed using the feature vectors extracted by our trained model exhibit greater alignment with the ground truth than those generated by its classical counterpart, as evidenced by the AMI scores, indicating that our model produces more structured and meaningful feature vectors for classification. Similarly, the processed images generated by our model generally contribute more significantly to clustering performance as well. Therefore, compared to MLTCNN, our model yields more critical features for classification throughout the pipeline.

Note that there are cases where the processed images generated by MLTCNN outperform those produced by our approach in terms of clustering performance. However, even in such cases, the feature vector extracted by our model demonstrates superiority over the features from the MLTCNN model, effectively compensating for the representation disadvantages of the processed image. This observation highlights the validity and effectiveness of the quantum component in our model for extracting features that contribute to improved classification performance.

\section{Conclusion and Future Work}\label{conclusion}

In this study, we present a hybrid quantum-classical neural network (MLTQNN) that leverages multitask learning to enable efficient quantum data encoding. Additionally, the model utilizes a proposed location weight module together with quantum convolution operations to extract features from EO data for classification tasks. \F{To evaluate the validity of our proposed model, we utilized multiple EO datasets, and the experimental results demonstrated its effectiveness.} 

Additionally, we explored the generalizability of our approach and further examined the underlying factors contributing to this property using various techniques. Our investigation suggests that our hybrid model, even when trained with a limited number of samples, can extract more robust and significant features compared to its classical counterpart, implying the generalizability advantages of quantum machine learning.

Regardless, future research could explore the following directions: 1) developing more suitable and efficient encoding approaches for EO data from different modalities; 2) further investigating the potential of QML to address EO challenges related to generalizability, such as domain shifts.

% ---- Bibliography ----
\bibliographystyle{ieeetr}
\bibliography{reference}

%\printbibliography

\end{document}